\documentclass{article}

\PassOptionsToPackage{numbers}{natbib}
\usepackage[preprint]{neurips_2025}

\usepackage[utf8]{inputenc} 
\usepackage[T1]{fontenc}    
\usepackage{hyperref}       
\usepackage{url}            
\usepackage{booktabs}       
\usepackage{amsfonts}       
\usepackage{nicefrac}       
\usepackage{microtype}      
\usepackage{xcolor}         
\usepackage{graphicx}
\usepackage{amsmath}       
\usepackage{algorithm}
\usepackage{algpseudocode}
\usepackage{multirow}
\usepackage{xcolor}
\usepackage{comment}
\usepackage{multirow}
\title{Stream-CQSA: Exact Out-of-Memory Recovery for Attention}

\author{%
  Yiming Bian \quad Joshua M. Akey\\
  Lewis-Sigler Institute of Integrative Genomics\\
  Princeton University\\
  Princeton, NJ 08540 \\
  \texttt{\{yimingb, jakey\}@princeton.edu} \\
  }

\begin{document}

\maketitle

\begin{abstract}

Long-context large language models are limited not only by attention cost but also by out-of-memory (OOM) failures. A selected attention call may not fit in available device memory even when the kernel is optimized. Exact and approximate attention methods reduce memory use, but every fixed implementation still has a device-specific capacity boundary. We introduce Stream-CQSA, an attention-level OOM recovery framework based on CQS decomposition, derived from the theory of cyclic quorum sets (CQS). Stream-CQSA recursively partitions an infeasible attention call into independent subsequence tasks, executes each with a compatible inner kernel, and recomposes the local statistics to recover the full attention output. This recovery is exact relative to the wrapped attention kernel, whether that kernel is exact or approximate. Compared with FlashAttention-2, the major baseline, our native Stream-CQSA kernel improves 16-bit forward-output error relative to a dense float64 reference and matches 16-bit backward-gradient error where FlashAttention-2 fits in the GPU memory. At the longest feasible baseline length, it costs $1.5$--$1.9\times$ the forward runtime and $2.1$--$2.4\times$ the forward--backward runtime. Beyond that sequence length boundary, our method continues to return an output while FlashAttention-2 OOMs. Stream-CQSA is therefore not a faster attention method. Instead, it converts memory-capacity failure into a recoverable execution path by trading extra compute, host-device transfer, and recomposition for completion.

\end{abstract}

\section{Introduction}
Modern large language model (LLM) capabilities depend strongly on model size and context length~\cite{kaplan2020scaling}. Recent systems~\cite{claude,comanici2025gemini,guo2025deepseek,yang2025qwen3} support context windows from hundreds of thousands to one million tokens, yet long inputs still degrade model behavior and remain expensive to process. A central barrier is the quadratic memory cost of exact self-attention~\cite{vaswani2017attention}. In practical training and inference pipelines, GPU memory often becomes the limiting resource even when the attention kernel is optimized. Cached activations, temporary buffers, and intermediate tensors can create run-time memory pressure that is hard to estimate statically. An otherwise well-defined attention computation may therefore fail at execution time and terminate the entire run. This motivates an automatic recovery mechanism that is independent of any particular sequence length or fixed memory budget. 

\begin{figure}
  \centering
  \centerline{\includegraphics[width=0.4\linewidth]{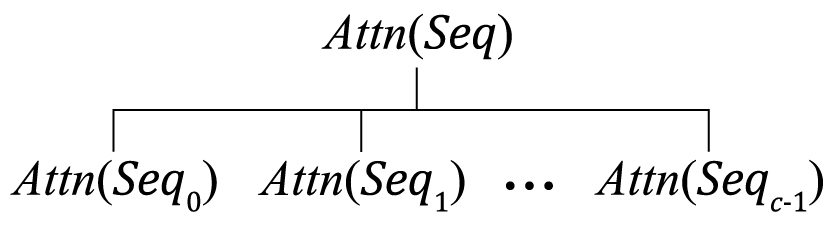}}
  \caption{CQS decomposition}
  \label{fig:cqs_divide}
\end{figure}

This work accordingly addresses attention-level OOM recovery, in which the target attention computation has already been selected but the corresponding call cannot complete within the available memory budget. We propose CQS decomposition (Figure~\ref{fig:cqs_divide}), an operation derived from cyclic quorum set (CQS) theory which partitions attention into independent subsequence computations whose local statistics recompose exactly in full-token coordinates. Stream-CQSA builds on this operation by increasing decomposition depth until the subproblems fit, executing them with a compatible inner kernel, and merging their statistics to recover the intended result. Existing attention methods can serve as inner kernels if they expose the required local statistics. Therefore, The objective of this work is neither to define a new efficient attention mechanism, nor to claim a speedup where the original computation already fits. It is to render an otherwise infeasible call executable.

\section{Prior work and positioning of this work}
Prior work improves long-context attention by reducing data movement, changing the attention rule, or distributing the computation. Exact methods such as FlashAttention~\cite{dao2022flashattention,dao2023flashattention} reduce memory movement and intermediate storage while preserving standard attention. Approximate methods reduce complexity by changing the computation, including sparse and blockwise attention~\cite{beltagy2020longformer,zaheer2020big,qiu2020blockwise,chen2023longlora}, dilated attention~\cite{ding2023longnet}, low-rank attention~\cite{wang2020linformer}, kernelized attention~\cite{kitaev2020reformer,katharopoulos2020transformers,choromanski2020rethinking}, and global-local or streaming mechanisms~\cite{xiao2023efficient}. Distributed methods shard the sequence across devices. Ring Self-Attention~\cite{li2023sequence} rotates keys and values around a device ring, Ring Attention~\cite{liu2024ringattention} extends this approach, Striped Attention~\cite{brandon2023striped} adapts it to causal masking, and DeepSpeed-Ulysses~\cite{jacobs2023deepspeed} and DistFlashAttn~\cite{li2023distflashattn} use different communication patterns. These methods improve, approximate, or distribute attention. Under a fixed device budget, each still has a capacity boundary.

Stream-CQSA instead recovers a computation that has already failed under a given budget. Out-of-core attention \cite{rabe2021self} is closest, but it only avoids materializing the quadratic attention matrix. $Q$, $K$, $V$, outputs, and gradients remain resident $\Theta(ND)$ terms, which account for $70$--$79\%$ of FlashAttention-2's peak memory in our measurements. Once these terms exceed the budget, smaller tiles cannot help. Sequence-parallel methods~\cite{li2023sequence,liu2024ringattention, brandon2023striped, jacobs2023deepspeed, li2023distflashattn} move the boundary outward by roughly the number of devices, but the degree of parallelism is fixed at launch and a shard that remains too large requires reconfiguration and restart. Stream-CQSA partitions the token set instead. A subproblem's working set shrinks with decomposition depth, every query--key pair is still computed exactly once, and subproblems communicate only during the final reduction. A failing subproblem can be decomposed further at run time (Section~\ref{sec:oom_guardrail}), and devices need not be identical. Stream-CQSA is complementary to the exact, approximate, and distributed attention families mentioned above. A method can serve as the inner attention kernels as long as it exposes recomposition statistics and supports the CQS mask. Relative to CQS-Attention \cite{bian2025cqs}, Stream-CQSA adds new features such as recursive decomposition, a pluggable inner-kernel interface, workload scheduling, and OOM guardrails.

\section{Stream cyclic quorum sets attention}
\label{sec:method}
Attention can be viewed as all pairwise token interactions. Stream-CQSA treats the selected attention call as the target and partitions it into subsequence computations whose retained interactions union exactly to the full set of token pairs. The main question is how many subsets are needed and which chunks each should contain.

Suppose the sequence is to be divided into $c$ subsequences while preserving all interactions. We first need to partition it into $c$ chunks, indexed from $0$ to $c-1$, thus $c(c-1)/2$ distinct off-diagonal chunk pairs. Each subsequence selects $l$ chunks and contributes $l(l-1)/2$ such pairs. Non-redundant coverage therefore requires $c\,l(l-1)/2=c(c-1)/2$, or $c=l(l-1)+1$. Therefore, valid $c$ values are $\{1, 3, 7, 13, 21, \dots\}$ for $l=\{1,2,3,4,5,\dots\}$. Essentially, CQS decomposition splits a complete graph ($K_c$) into $c$ smaller complete subgraphs ($K_l$) that every edge in $K_c$ is covered exactly once, corresponding to a Steiner system $S(2,l,c)$~\cite{stinson2008combinatorial} when it exists. 

The cases $l=1, c=1$ and $l=2, c=3$ are trivial. Therefore, we use the smallest non-trivial case $l=3, c=7$ throughout the discussion. The sequence is partitioned into $7$ chunks $(C_0, \dots, C_6)$, and each subsequence $i$ is constructed as
\begin{equation}
    Seq_i=\texttt{concat}(C_{(0+i)\text{ mod } 7},C_{(1+i)\text{ mod }7},C_{(3+i)\text{ mod }7})
\end{equation}
for $0 \le i \le 6$. For notational simplicity, we denote subsequences by their chunk indices. The construction is cyclic, with the base pattern $(0,1,3)$ defining the first subsequence. We refer to this pattern as the interest set ($\mathcal{I}$) for $c=7$~\cite{bian2021efficient}, so that $|\mathcal{I}|=l$. We visualize the full coverage under this construction in Figure~\ref{fig:fwd}, and further discussion on CQS interest set is given in Appendix~\ref{appendix:CQS}.

\subsection{Forward pass}
\label{sec:fwd}

For each subsequence $i$, we gather $Q_i$, $K_i$, and $V_i$ from the full-sequence tensors $Q$, $K$, and $V$. CQSA~\cite{bian2025cqs} then performs the following computations:
\begin{equation}
    R_i=\alpha(Q_iK_i^\top) \quad P_i=\text{exp}(R_i)\odot M_i \quad \mathrm{Num}_i=P_iV_i \quad \mathrm{Den}_i=\mathrm{row\_sum}(P_i)
\end{equation}
where $\alpha=1/\sqrt{D}$, $D$ is head dimension, $M_i$ is the CQS binary mask (Section~\ref{sec:cqs_mask}), and $\mathrm{Num}_i,\mathrm{Den}_i$ correspond to the numerator and denominator of the softmax computation.

CQSA then scatters local statistics into full-token accumulators and normalizes as
\begin{equation}
    \mathrm{Num}=\sum_i\text{scatter}(\mathrm{Num}_i) \quad \mathrm{Den}=\sum_i\text{scatter}(\mathrm{Den}_i) \quad O=\mathrm{Num} \oslash\mathrm{Den}
\end{equation}
where $\oslash$ denotes token-wise division broadcast along the head dimension ($D$).

\begin{figure}
  \centering
  \centerline{\includegraphics[width=1\linewidth]{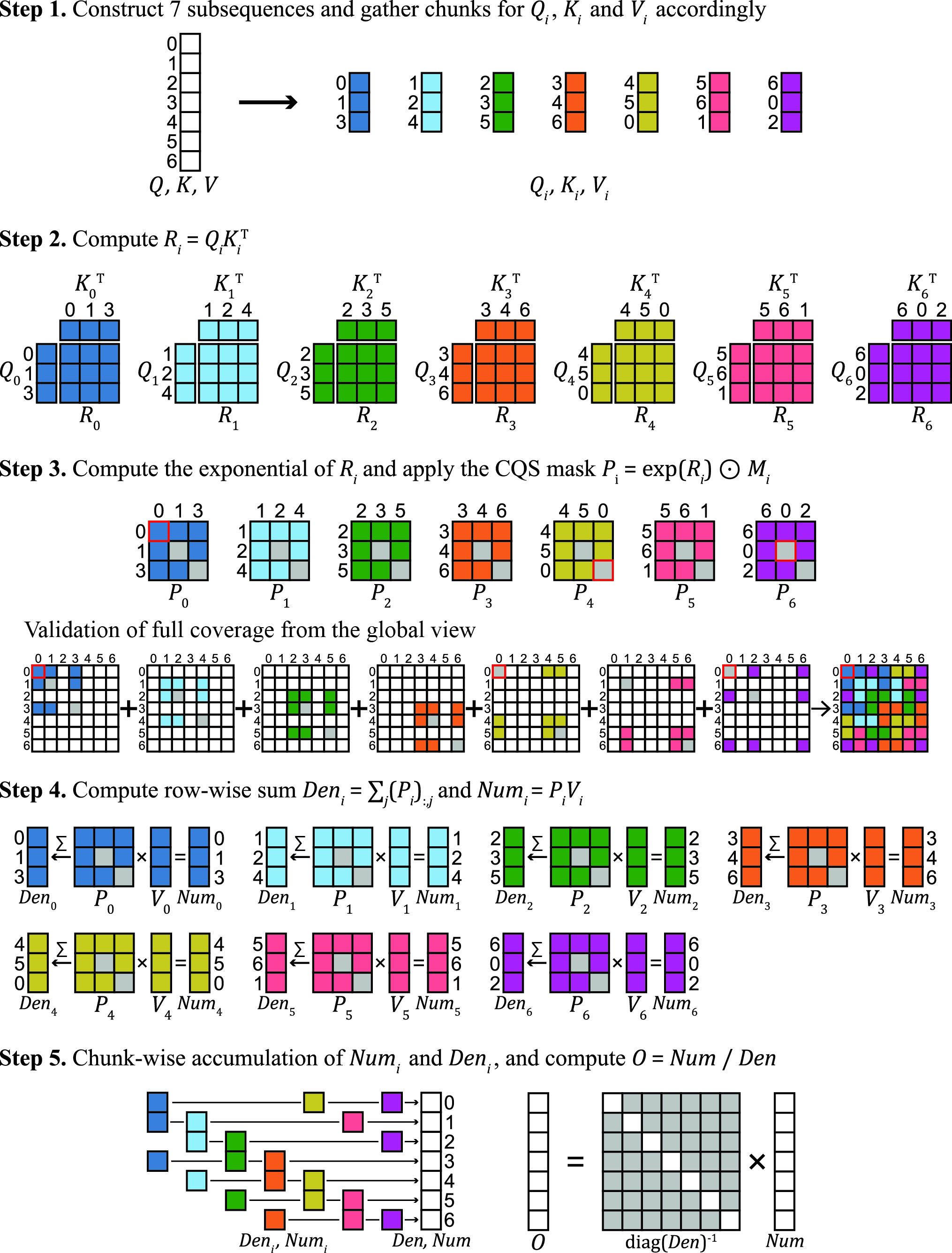}}
  \caption{CQSA forward pass with $c=7$ and $\mathcal{I}=(0,1,3)$. Step 3 highlights chunk pair $(0,0)$, which appears in subsequences $0$, $4$, and $6$. The CQS mask $M_i$ keeps exactly one contribution and masks the rest. The global attention view confirms that every chunk pair is covered once. In step 5, $\mathrm{Num}$ and $\mathrm{Den}$ are accumulated the same way but have dimensions $N\times D$ and $N\times 1$, respectively.}
  \label{fig:fwd}
\end{figure}

The forward pass is summarized in Algorithm~\ref{algo:fwd}. Figure~\ref{fig:fwd} provides a step-by-step illustration for $c=7$ and $\mathcal{I}=(0,1,3)$, explicitly verifying full coverage of the attention matrix $P$.

\subsection{Backward pass}
The backward pass follows the same decomposition. Let $\textbf{d}O$ denote the upstream gradient. Since $O=\mathrm{Num}\oslash \mathrm{Den}$, we have
\begin{equation}
    \textbf{d}\mathrm{Num}=\textbf{d}O\oslash \mathrm{Den} \quad \quad \textbf{d}\mathrm{Den}=-\frac{\langle \textbf{d}O,\mathrm{Num} \rangle _D}{\mathrm{Den}^2}
\end{equation}
where $\langle\cdot,\cdot\rangle_D$ denotes the inner product on the head dimension. 

Using the same $(c,\mathcal{I})$, we gather $\textbf{d}\mathrm{Num}_i$ and $\textbf{d}\mathrm{Den}_i$ for each subsequence and compute
\begin{equation}
\begin{split}
    \textbf{d}V_i=P_i^ \top \textbf{d}\mathrm{Num}_i \quad \textbf{d}P_i=\textbf{d}\mathrm{Num}_iV_i^ \top+\textbf{d}\mathrm{Den}_i \mathbf{1}^ \top \\ \textbf{d}R_i=\textbf{d}P_i \odot P_i \quad \textbf{d}Q_i=\alpha \textbf{d}R_iK_i \quad \textbf{d}K_i=\alpha \textbf{d}R_i^ \top Q_i
\end{split}
\end{equation}
Finally, $\textbf{d}Q_i$, $\textbf{d}K_i$, and $\textbf{d}V_i$ are scattered into full-sequence coordinates exactly as in the forward pass
\begin{equation}
    \textbf{d}Q=\sum_i \text{scatter}(\textbf{d}Q_i) \quad \textbf{d}K=\sum_i \text{scatter}(\textbf{d}K_i) \quad \textbf{d}V=\sum_i \text{scatter}(\textbf{d}V_i)
\end{equation}
The backward pass is stated formally in Algorithm~\ref{algo:bwd}, with an illustration in Figure~\ref{fig:bwd} for $c=7$ and $\mathcal{I}=(0,1,3)$. A full chain-rule derivation is given in Appendix~\ref{appendix:bwd}.

\subsection{CQS masking}
\label{sec:cqs_mask}

Subsequences constructed by CQS decomposition exhibit structured overlap. This overlap guarantees full coverage of all token interactions but also introduces redundancy: since each chunk appears in $l$ subsequences, interactions within diagonal chunk pairs are computed $l$ times. Aggregating them directly would yield an incorrect result. Correctness therefore requires that each subsequence be assigned a predetermined responsibility over the attention matrix, with redundant contributions removed, which introduces the CQS mask. The masking rule is simple: subsequence $i$ is responsible only for the diagonal chunk pair $(i,i)$. For example, subsequence $0$ with $\mathcal{I}=(0,1,3)$ contains $(0,0)$, $(1,1)$, and $(3,3)$. It retains only $(0,0)$ because the other two belong to subsequences $1$ and $3$.

\begin{figure}
  \centering
  \centerline{\includegraphics[width=1\linewidth]{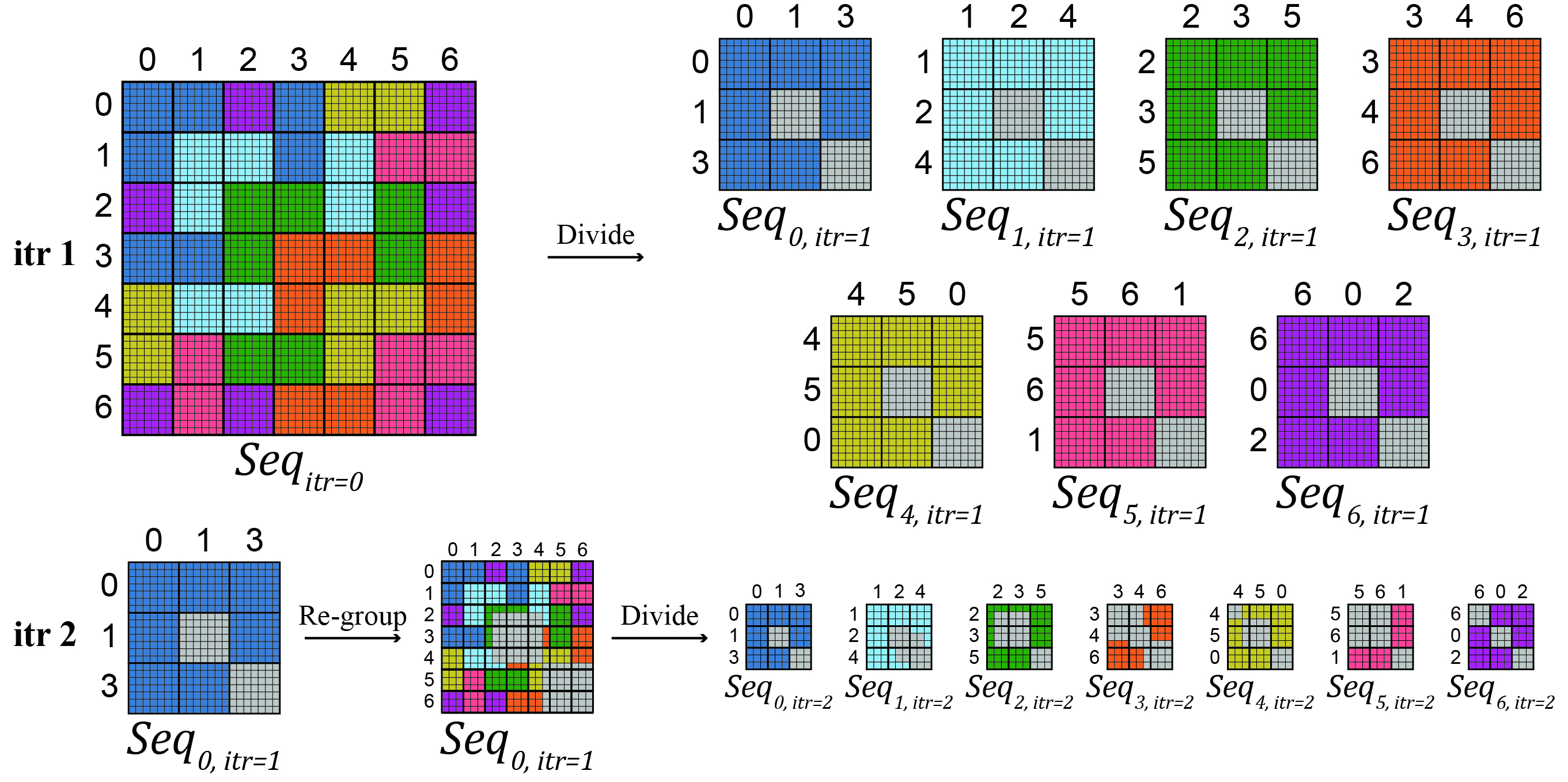}}
  \caption{CQS masking on $R = QK^\top$ with $N=49$, $c=7$, and $\mathcal{I}=(0,1,3)$. Depth 1 forms $21$-token subsequences and masks repeated diagonal regions. Depth 2 repartitions each subsequence into chunks of size $3$, producing $9$-token subsequences with local and inherited masks. Only $Seq_{0,\texttt{itr}=1}$ is shown here. Other depth $1$ subsequences ($Seq_{i,\texttt{itr}=1}$) follow the same pattern.}
  \label{fig:cqs_mask}
\end{figure}

CQS decomposition can be applied recursively because each subsequence is itself a sequence. We write $\texttt{itr}$ for the decomposition depth. Depth $\texttt{itr}$ yields $c^{\texttt{itr}}$ subsequences. At depth $\texttt{itr}+1$, each subsequence applies the same diagonal CQS masking rule and inherits the masks from earlier depths, producing structured off-diagonal masks. Figure~\ref{fig:cqs_mask} shows CQS masks at depths $1$ and $2$ for $N=49$.

After masking, subsequence computations are non-redundant and independent. Stream-CQSA can therefore execute them in parallel, distribute them across devices, and schedule them on heterogeneous hardware (Section~\ref{sec:workload_sched}). The choice of $c$ is also flexible. Different valid values, such as those listed in Table~\ref{tab:interest_set}, may be used at each depth, affording finer control over subsequence length and thus closer utilization of the available memory.

\subsection{OOM guardrails}
\label{sec:oom_guardrail}

Subsequence independence is also what makes recovery possible. If a subsequence still exceeds the memory budget, CQS decomposition can split it again into smaller independent subproblems whose local statistics recompose in the same way. This recursion continues until every subproblem fits, subject to the minimal buffers required by the inner attention kernel and the host-side storage required for recomposition. The guardrail built on this property rests on a single rule: whenever the current configuration exhausts memory, increase the decomposition depth ($\texttt{itr}$) by one. Since it stops at the first depth that does not fail, it returns the smallest feasible depth by construction.

The second parameter, $\texttt{n\_cap}$, bounds the number of subsequences resident on the GPU at once and adjusts within a depth before the depth itself changes. On a failure the offending subsequence is evicted and returned to the scheduling queue and $\texttt{n\_cap}$ is reduced, which lowers memory pressure without altering the decomposition. Depth changes only when $\texttt{n\_cap}$ reaches either limit of its range. If $\texttt{n\_cap}=1$ OOMs, no further reduction is available and the depth increases. If $\texttt{n\_cap}=c^{\texttt{itr}}$ and headroom remains, the decomposition is finer than necessary and the depth decreases. After any depth change, calibration re-establishes $\texttt{n\_cap}$. The two limits of $\texttt{n\_cap}$ thus bracket the depth from below and above, and $\texttt{n\_cap}$ is re-established by a calibration pass after any change of depth. We stress that $\texttt{n\_cap}$ bounds residency rather than concurrency. How many resident subsequences execute simultaneously is a property of the inner kernel.

\subsection{Workload scheduling}
\label{sec:workload_sched}

Once the guardrail has established a feasible decomposition, the remaining question is placement. A CQS subsequence computes independently of every other and the local statistics recompose in any order, so a monolithic attention call becomes a set of schedulable tasks. Any assignment of these tasks to devices and to time that respects the memory limit of each device yields the same exact result. Correctness therefore constrains the decomposition but not the schedule.

\begin{figure}
  \centering
  \centerline{\includegraphics[width=1\linewidth]{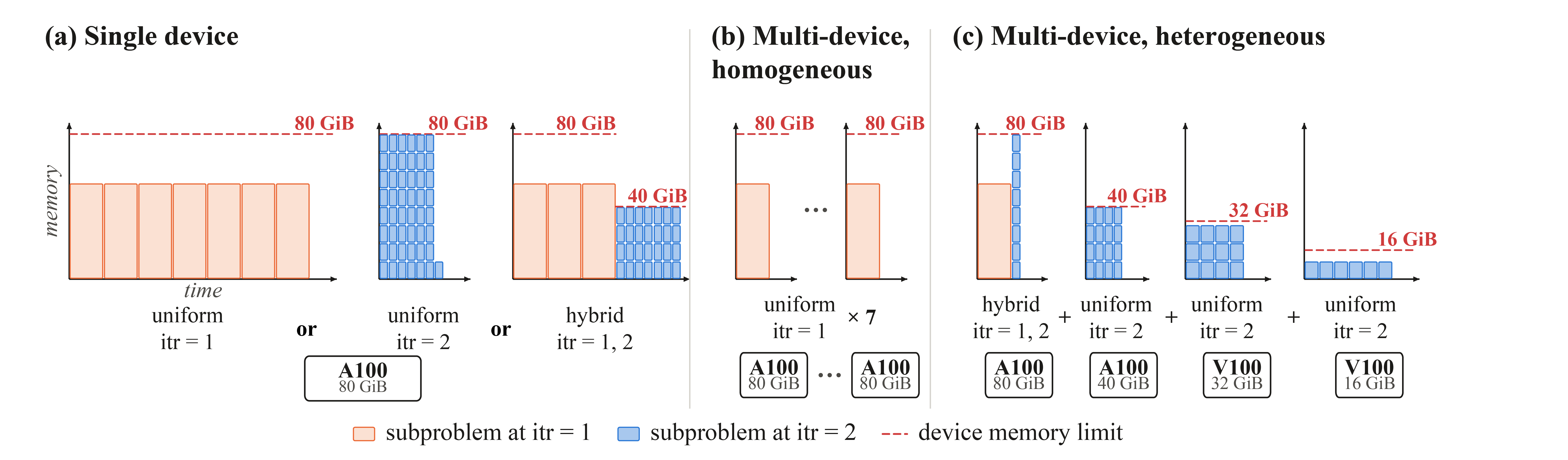}}
  \caption{Schedules admitted by the decomposition. Both axes are absolute and shared across the panels so that blocks are directly comparable in the figure. The workload is hypothetical and the per-subproblem costs are chosen only so that the schedules may be drawn to scale, with the V100 assumed $1.6\times$ slower than the A100. \textbf{(a)} Three schedules for a single device, including uniform schedules at either granularity and a hybrid that mixes them. \textbf{(b)} Identical devices, with one subproblem on each. \textbf{(c)} Devices of different capacities, each scheduling under its own limit. A given subproblem has the same height but is wider on the slower device.}
\label{fig:schedule}
\end{figure}

Figure~\ref{fig:schedule} illustrates this flexibility. On a single device, a further level of decomposition produces more subproblems that are individually cheaper, so the total workload rises. It also reduces the memory each subproblem requires and allows several to be resident at once, so the schedule may complete sooner despite performing more work. A hybrid schedule mixes the two depths. It arises most often not by design but from a change in the memory available during execution. The OOM guardrail decomposes the remaining branches further while those already completed stay at the coarser depth. Across devices, independence removes communication from the critical path, no exchange being required until the final summation. Where the devices differ, two properties govern the assignment. The feasible depth is a property of the device, because a subproblem that does not fit compels further decomposition, which is what allows a smaller device to participate. The per-subproblem cost is instead a property of the hardware, unchanged in memory but greater in time on a slower device. A scheduler must account for both. All experiments in this work use the single-device uniform schedule (Figure~\ref{fig:schedule}a). The other schedules show what the decomposition permits, not configurations we benchmark. Scheduler design is an important problem and is left to future work.

\section{Experiments}
\label{sec:results}

The following evaluation addresses three claims. The first is \textbf{correctness under precision}: decomposition and recomposition should not change the attention output beyond dtype-level error, in either forwand or backward pass at supported precisions. The second is \textbf{the cost where the unwrapped monolithic path fits}: Stream-CQSA is expected to be slower and we quantify that overhead. The third is \textbf{recovery}: beyond the baseline OOM sequence length boundary, Stream-CQSA should return a result, with the decomposition depth increasing as the memory budget tightens.

\subsection{Setup}
\label{sec:exp_setup}

All measurements use causal attention with $B=1$, $H=8$, $D=64$ on a single NVIDIA A100 80\,GiB, PyTorch 2.10.0+cu130 and \texttt{flash-attn} 2.8.3. CQS decomposition uses $c=7$ and $\mathcal{I}=(0,1,3)$, and every Stream-CQSA measurement runs on the native kernel of Appendix~\ref{sec:native_kernel}, which is modified based on FlashAttention-2 to evaluate the CQS mask and to skip the tiles that mask leaves empty. Synthetic $Q,K,V$ and the upstream gradient $\mathbf{d}O$ are drawn from a zero-mean unit-variance Gaussian with fixed seeds and generated once per $(N,\text{dtype})$, so every method sees identical inputs at each point. We test fp16 and bf16 because the FlashAttention tensor-core path does not accept fp32, and the output, the softmax statistics and the merge are computed in fp32.

\paragraph{Baselines.} The main experiments compare exact attention implementations only and one compatible approximate-attention case is demonstrated separately in Appendix~\ref{sec:appendix_experiments}. \emph{SDPA} is PyTorch's default dispatcher, which by default selects a FlashAttention backend on this hardware. \emph{FlashAttention-2} is the reference \texttt{flash-attn} package~\citep{dao2022flashattention,dao2023flashattention}. We report both because they differ by $1$--$8\%$ across the measured range, due to dispatch overhead and bundled-build differences. The third baseline is \emph{SDPA (mem-eff.)}, which pins PyTorch's memory-efficient backend and is the only baseline whose internals differ.

\paragraph{Stream-CQSA configurations.} We test three configurations that vary in two aspects. The first is the decomposition depth $\texttt{itr}$, which determines the number of subproblems. The second is the residency of the fp32 accumulator, which is held either on the device or in host memory. In all three, the full-length $Q$, $K$ and $V$ remain in host memory for the whole call, and only the gathered subsequences $Q_i$, $K_i$, $V_i$ of the subproblems currently in flight are copied to the device. Two fixed-depth configurations are $\texttt{itr}=1$ and $\texttt{itr}=2$, both with the accumulator on the device and the OOM guardrail disabled on purpose. These two variants aim to demonstrate the time-memory tradeoff and they are not recommended in practice. The third is $\texttt{itr}=\textsc{auto}^{*}$ with $\min(\texttt{itr})=1$, the accumulator in host memory and OOM guardrail engaged. So the depth reported is either $1$ or the smallest at which no subproblem exhausts memory. The star distinguishes it from the $\texttt{itr}=\textsc{auto}$ of the shipped package, which begins at $\texttt{itr}=0$ and returns a single monolithic call when possible to avoid the decomposition cost.

\paragraph{Metrics.} Peak device memory (GiB) is \texttt{torch.cuda.max\_memory\_allocated()} over the call, with the allocator drained between measurements. Wall-clock time (s) is the median over repeats. We report two timing scopes. A forward timing measures only the forward attention call and corresponds to prefill. A forward--backward timing measures the full training-step attention cost by running the forward call followed by gradient computation. We use this scope because the gradient is not a standalone computation. It depends on intermediates produced by the same forward call, including the output, softmax statistics, and for Stream-CQSA the global log-sum-exp produced by the decomposed forward. Timing only the gradient would either reuse intermediates created outside the measured region or charge one configuration for intermediates prepared by another. Neither case is a cost a user pays in isolation. This forward--backward measurement gives every method the same scope and keeps ratios comparable. Per-stage breakdowns use CUDA events. When several subproblems are in flight, event spans include time queued behind other streams, so they show where work is issued rather than a strict runtime partition.

\subsection{Output preservation under precision}
\label{sec:exp_accuracy}

All accuracy figures in the forward and backward pass reported in Table~\ref{tab:accuracy} are relative errors against a \textbf{float64} reference computed from the same inputs. A dense float64 is $O(N^2)$ and we fix the sequence length to $N=8\,192$ so that a dense backward reference remains tractable. We average the difference over 10 random draws. In the forward pass, the decomposed path is more accurate than every baseline. The gain is because Stream-CQSA (denoted as $\mathcal{S}$) accumulates in fp32 and returns fp32, whereas the baselines round their output back to the 16-bit input dtype. In the backward it is comparable, differing from the baselines by at most $0.3\%$ at either precision. Because there is no fp32-to-16-bit rounding happens in the backward for all implementations. Overall, the decomposition-recomposition mechanism of Stream-CQSA preserves the attention function to the precision of the operands.

\begin{table}[t]
\centering\small
\caption{Relative error against the float64 reference. Backward error by gradient tensor is given in Table~\ref{tab:accuracy_grad}. $\mathcal{S}$ denotes Stream-CQSA. The $\texttt{itr}=0$ row is the monolithic or \emph{undecomposed} path, which is what the shipped package $\texttt{itr}=\textsc{auto}$ selects at $N=8192$.}
\label{tab:accuracy}
\begin{tabular}{lcccc}
\toprule
\multirow{2}{*}{Method} & \multicolumn{2}{c}{Forward} & \multicolumn{2}{c}{Backward} \\
\cmidrule(lr){2-3}\cmidrule(lr){4-5}
 & fp16 & bf16 & fp16 & bf16 \\
\midrule
SDPA & $2.689\!\times\!10^{-4}$ & $2.168\!\times\!10^{-3}$ & $3.075\!\times\!10^{-4}$ & $2.458\!\times\!10^{-3}$ \\
SDPA (mem-eff.) & $2.685\!\times\!10^{-4}$ & $2.159\!\times\!10^{-3}$ & $3.739\!\times\!10^{-4}$ & $2.980\!\times\!10^{-3}$ \\
FlashAttention-2 & $2.689\!\times\!10^{-4}$ & $2.168\!\times\!10^{-3}$ & $3.075\!\times\!10^{-4}$ & $2.458\!\times\!10^{-3}$ \\
\midrule
$\mathcal{S}$ \textsc{itr}=0 (undecomposed) & $2.689\!\times\!10^{-4}$ & $2.168\!\times\!10^{-3}$ & $3.075\!\times\!10^{-4}$ & $2.458\!\times\!10^{-3}$ \\
$\mathcal{S}$ \textsc{itr}=1 & $1.705\!\times\!10^{-4}$ & $1.375\!\times\!10^{-3}$ & $3.080\!\times\!10^{-4}$ & $2.462\!\times\!10^{-3}$ \\
$\mathcal{S}$ \textsc{itr}=2 & $1.681\!\times\!10^{-4}$ & $1.356\!\times\!10^{-3}$ & $3.085\!\times\!10^{-4}$ & $2.465\!\times\!10^{-3}$ \\
\bottomrule
\end{tabular}
\end{table}

\begin{figure}[t]
\centering
\includegraphics[width=0.8\linewidth]{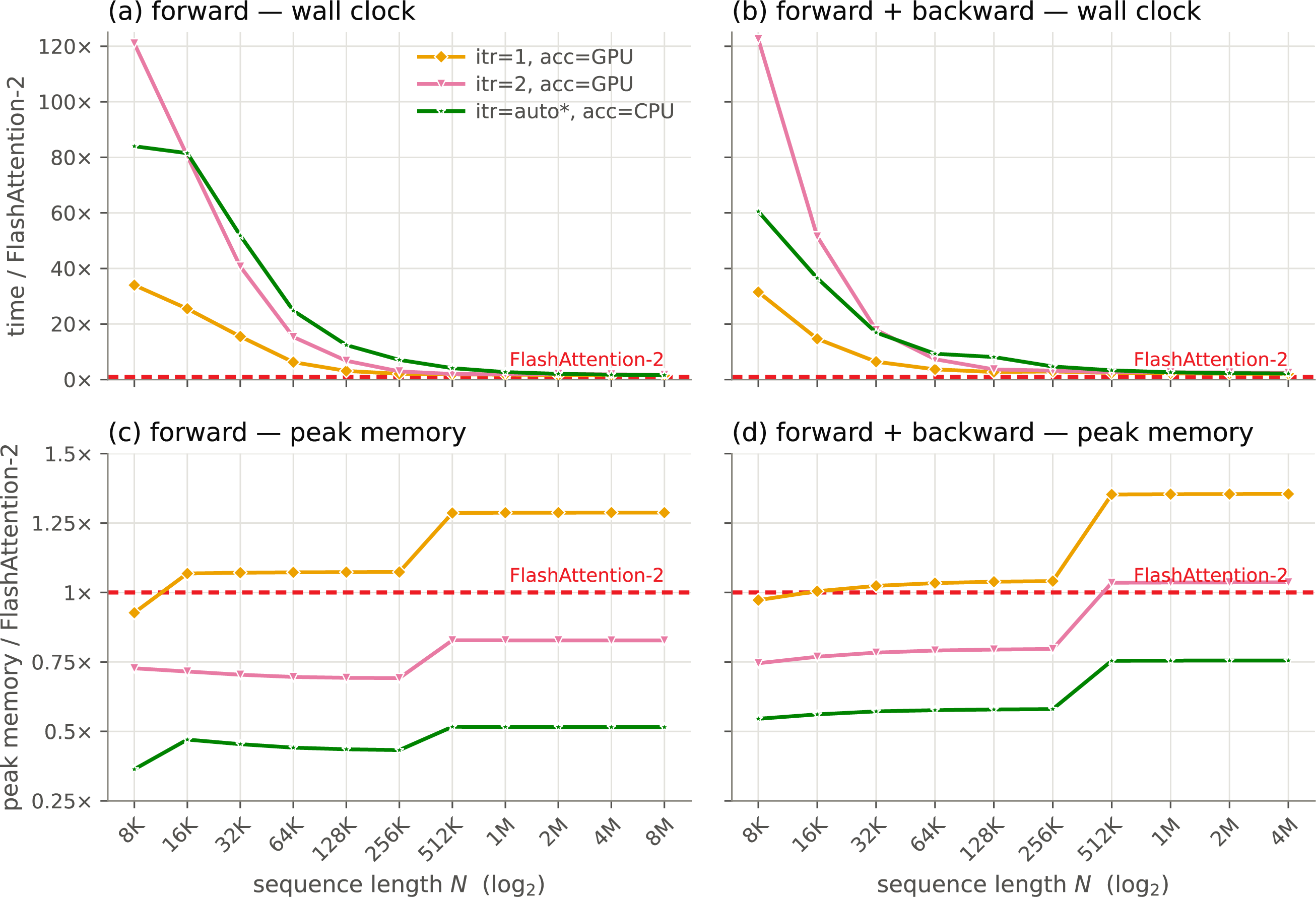}
\caption{Cost of three Stream-CQSA configurations relative to FlashAttention-2. The forward panels run up to $8\text{M}$ and the forward--backward panels to $4\text{M}$. The dashed line marks parity with FlashAttention-2. The underlying measurements are given in Tables~\ref{tab:supp_fwd} and~\ref{tab:supp_bwd}.}
\label{fig:overhead_ratio}
\end{figure}

\subsection{Cost when the monolithic call fits}
\label{sec:exp_overhead}

This subsection and Section~\ref{sec:exp_recovery} use one sweep over twelve sequence lengths from $8\text{K}$ to $16\text{M}$ in powers of two. Here we report the range where FlashAttention-2 completes, with ratios defined up to $8\text{M}$ for forward and $4\text{M}$ for forward--backward. Figure~\ref{fig:overhead_ratio} shows that overhead is large at short lengths and falls quickly. In the forward pass, $\texttt{itr}=\textsc{auto}^*$ costs $84\times$ at $8\text{K}$, $2.68\times$ at $1\text{M}$, $1.76\times$ at $4\text{M}$, and $1.59\times$ at $8\text{M}$. Fixed-depth configurations follow similar curves. Forward--backward timing settles slightly higher, at $2.17\times$ at $4\text{M}$. The decline occurs because gather, staging, and recomposition costs scale mainly with the number of subproblems, while attention arithmetic grows with $N^2$. Thus overhead is worst where recovery is least useful and it becomes smallest near the memory boundary. Memory shows the opposite tradeoff. From $N=512\text{K}$ upward with the accumulator on device, Stream-CQSA converges to $1.29\times$ forward and $1.35\times$ forward--backward memory at $\texttt{itr}=1$, and to $0.83\times$ and $1.04\times$ at $\texttt{itr}=2$. Moving the accumulator to host is the only configuration below baseline memory in both passes, at $0.52\times$ forward and $0.76\times$ forward--backward. Stream-CQSA therefore pays for subsequence construction, transfers, repeated kernel launches, and recomposition. Therefore, it is an OOM recovery mechanism, not the preferred path when a monolithic call fits.

\subsection{Recovery beyond the memory boundary}
\label{sec:exp_recovery}

\begin{figure}[t]
\centering
\includegraphics[width=\linewidth]{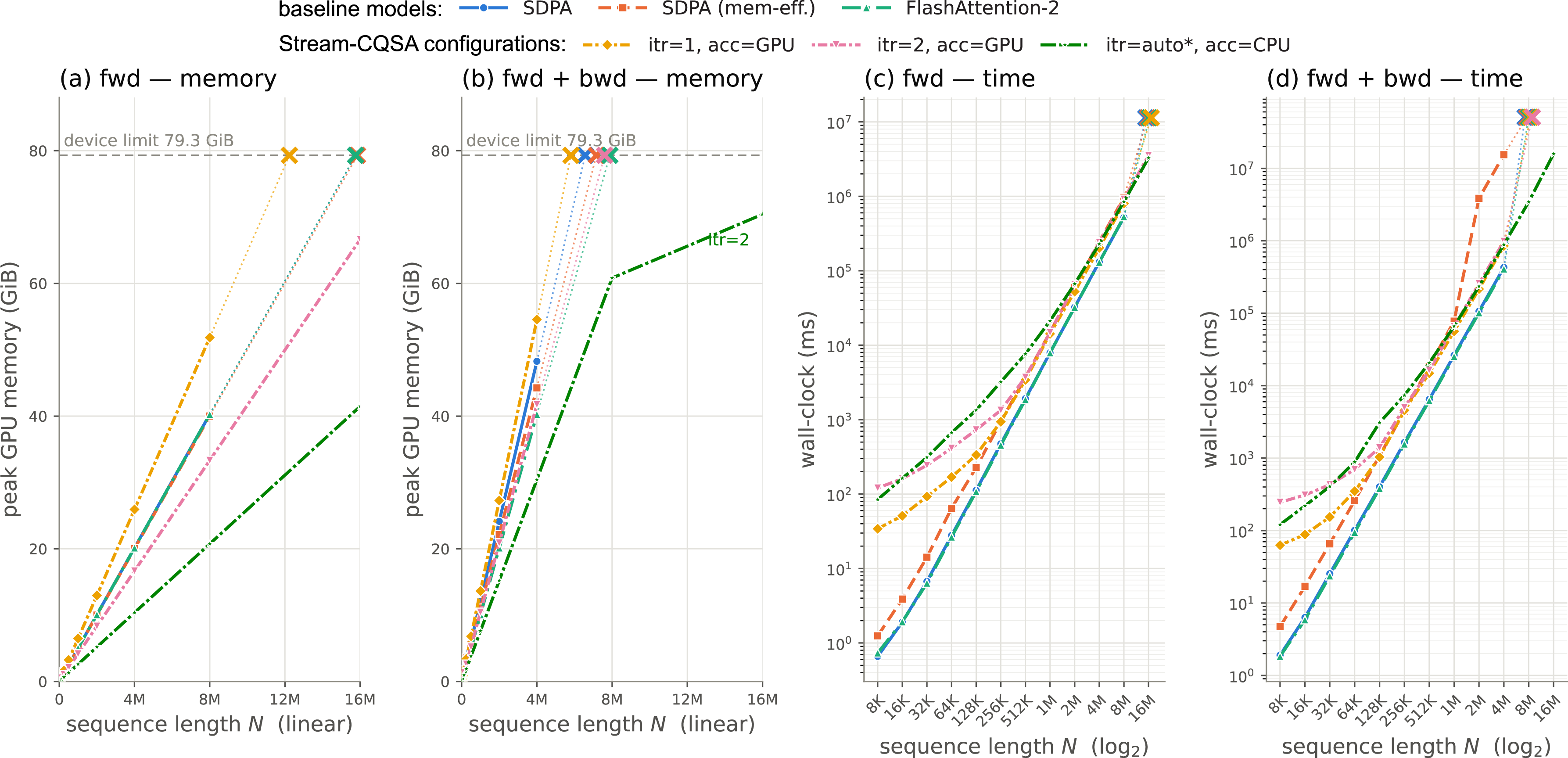}
\caption{Peak device memory and wall-clock time versus sequence length}
\label{fig:mem_time_fp16}
\end{figure}

We next increase $N$ to $16\text{M}$ (Figure~\ref{fig:mem_time_fp16}), beyond the baseline GPU-memory boundary. Peak memory is linear in $N$, so each curve reaches the $79.3$ GiB device limit at a predictable length. All baselines and Stream-CQSA $\texttt{itr}=1$ fail at $16\text{M}$ forward and $8\text{M}$ forward--backward. Fixed $\texttt{itr}=2$ completes $16\text{M}$ forward but fails at $8\text{M}$ forward--backward. Stream-CQSA $\texttt{itr}=\textsc{auto}^*$ with the OOM guardrail completes all measured lengths in both directions. Table~\ref{tab:recovery} summarizes the performance at the memory boundary. The baselines fail first, $\textsc{acc}$=GPU survives further up to the maximum capacity of the chosen depth, and $\textsc{acc}$=CPU never fails and has a slower memory growth rate because the OOM guardrail is engaged and the accumulator is relocated to host memory.

\begin{table}[t]
\centering\small
\caption{Memory boundary and recovery. Entries are time (s) / peak memory (GiB). The final row shows the depth reached by $\texttt{itr}=\textsc{auto}^{*}$. A star marks a length where the shipped $\texttt{itr}=\textsc{auto}$ would execute a monolithic call (i.e., $\texttt{itr}=0$).}
\label{tab:recovery}
\resizebox{\linewidth}{!}{%
\begin{tabular}{lcccccc}
\toprule
\multirow{2}{*}{Method} & \multicolumn{3}{c}{Forward} & \multicolumn{3}{c}{Forward--Backward} \\
\cmidrule(lr){2-4}\cmidrule(lr){5-7}
& $4\text{M}$ & $8\text{M}$ & $16\text{M}$ & $4\text{M}$ & $8\text{M}$ & $16\text{M}$ \\
\midrule
SDPA             & $131$ / $20.1$ & $520$ / $40.3$ & \textsc{oom} & $429$ / $48.3$ & \textsc{oom} & \textsc{oom} \\
SDPA (mem-eff.)  & $245$ / $20.0$ & $966$ / $40.0$ & \textsc{oom} & $15421$ / $44.3$ & \textsc{oom} & \textsc{oom} \\
FlashAttention-2 & $130$ / $20.1$ & $532$ / $40.3$ & \textsc{oom} & $407$ / $40.3$ & \textsc{oom} & \textsc{oom} \\
\midrule
$\mathcal{S}$, \textsc{itr}=1, \textsc{acc}=GPU & $203$ / $25.9$ & $802$ / $51.8$ & \textsc{oom} & $847$ / $54.5$ & \textsc{oom} & \textsc{oom} \\
$\mathcal{S}$, \textsc{itr}=2, \textsc{acc}=GPU & $240$ / $16.7$ & $914$ / $33.3$ & $3544$ / $66.6$ & $983$ / $41.7$ & \textsc{oom} & \textsc{oom} \\
$\mathcal{S}$, \textsc{itr}=\textsc{auto}$^{*}$, \textsc{acc}=CPU & $228$ / $10.4$ & $846$ / $20.7$ & $3277$ / $41.5$ & $885$ / $30.4$ & $3437$ / $60.8$ & $15847$ / $70.4$ \\
\midrule
depth $\texttt{itr}$ reached & $1^{*}$ & $1^{*}$ & $1$ & $1^{*}$ & $1$ & $2$ \\
\bottomrule
\end{tabular}}
\end{table}

One table entry of PyTorch's memory-efficient forward--backward path takes $15\,421$ s at $4\text{M}$, versus $407$ s for FlashAttention-2. Below $1\text{M}$, the same ratio is only $2.8$--$3.1\times$. The jump occurs between $1\text{M}$ and $2\text{M}$, where runtime grows by $49.6\times$ instead of the expected quadratic $4\times$, before quadratic scaling resumes from $2\text{M}$ to $4\text{M}$ (Figure~\ref{fig:mem_time_fp16}d). The cause is a CUDA grid-dimension limit. That backend tiles queries into $32$-row blocks mapped to a grid dimension capped at $65\,535$, creating a boundary at $65\,535\times32=2\,097\,120$ tokens. In a targeted check of the backward kernel, that backend took $36.5$ s at $N=2\,097\,088$ ($65\,534$ blocks) but $4250$ s at $N=2\,097\,120$ ($65\,535$ blocks), while FlashAttention-2 stayed around $13$ s in both cases. This explains the wall-clock time leap from $N=1$M ($32\,768$ blocks) to $N=2$M ($65\,536$ blocks).

\begin{figure}[t]
\centering
\includegraphics[width=\linewidth]{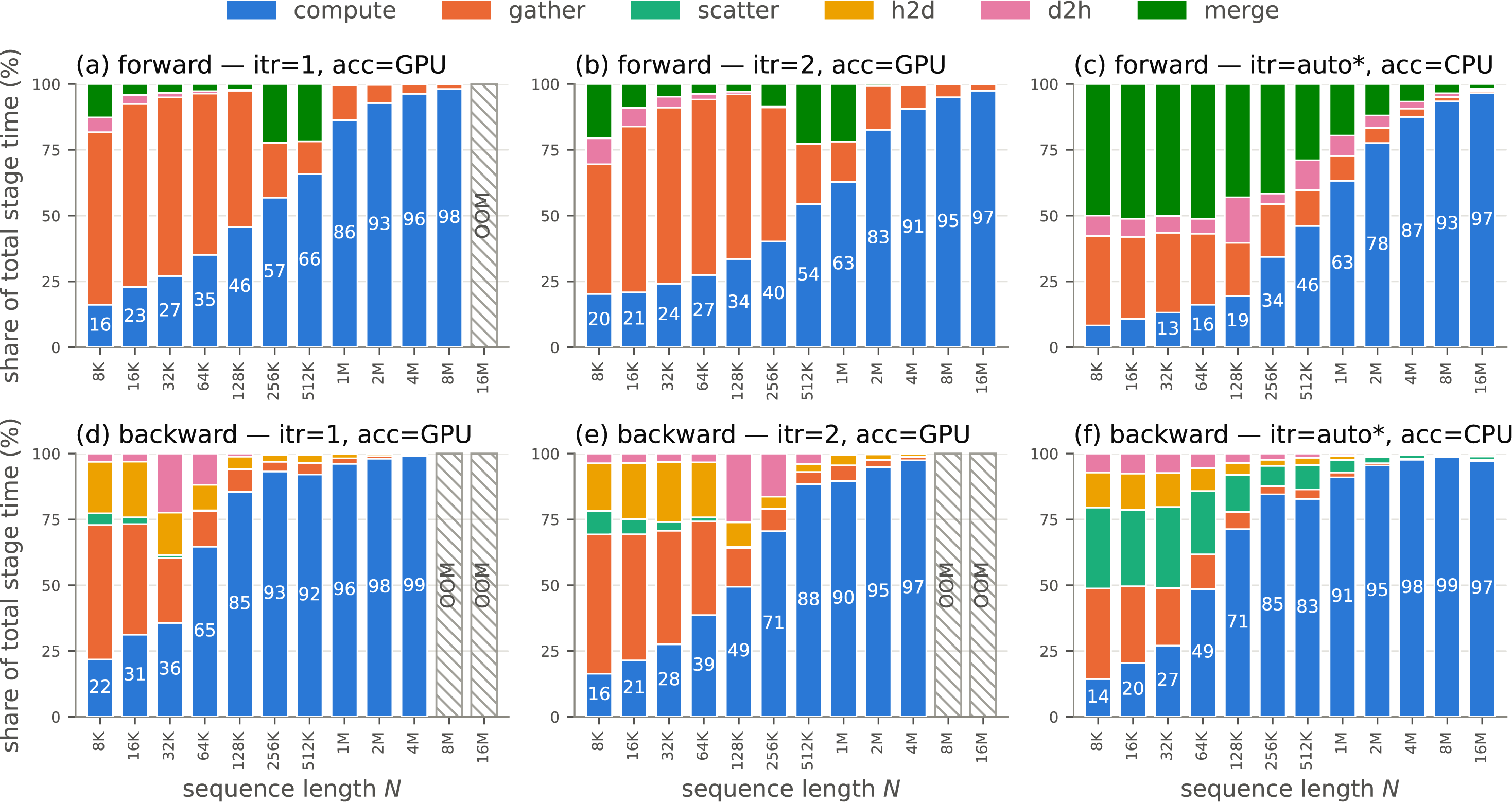}
\caption{Stage breakdown for three Stream-CQSA configurations. Stage totals come from CUDA events summed over subsequences and may exceed wall-clock time when streams overlap. The backward panels cover the backward pass alone, unlike forward--backward wall-clock and memory measurements elsewhere, which time a forward pass followed by gradient computation.}
\label{fig:stages}
\end{figure}

Figure~\ref{fig:stages} profiles three Stream-CQSA configurations by stage. Figure~\ref{fig:supp_stages_abs} shows the same measurements in absolute time on log-log axes. We report six stages. \texttt{gather} assembles subsequence $Q_i,K_i,V_i$ tensors, \texttt{scatter} returns local results to full-token coordinates, \texttt{h2d} and \texttt{d2h} transfer data, \texttt{compute} runs attention, and \texttt{merge} recomposes max-shifted partial statistics. The costs separate by order. \texttt{compute} is quadratic, while gather, scatter, and transfers are linear in $N$. Accumulator residency determines how much linear work remains. With a device accumulator, partial results are merged in place and \texttt{merge}/\texttt{d2h} are nearly constant. With a host accumulator, each subproblem's fp32 statistics cross the bus and merge on the host, which are both linear costs. While the sequence becomes longer, linear work measured against the quadratic work becomes negligible and the attention computation ($\texttt{compute}$) dominates the breakdown. For $\texttt{itr}=\textsc{auto}^{*}$ with $\textsc{acc}$=CPU, \texttt{compute} accounts for $96.5\%$ of total stage time in the forward panel and $97.2\%$ in the backward panel at $N=16\text{M}$. Stage totals sum CUDA-event spans across subsequences. They can exceed wall-clock time when streams overlap. At short lengths the sum reaches $3.18\times$ elapsed time. From $N=2\text{M}$ onward, only one subsequence is resident, and stage sum and wall clock differ by at most $3\%$. For example, they differ by $17$ s out of $3\,248$ s at $16\text{M}$. Therefore, in the very long sequence regime where a monolithic call fails and Stream-CQSA is designed for, the attention computation dominates the runtime and extra cost is almost negligible.

\section{Limitations and future work}
\label{sec:limitations}

This work addresses attention-level out-of-memory recovery through a decomposition-recomposition mechanism. The evaluation shows that Stream-CQSA preserves the attention output to within the precision of its operands and continues to return results beyond the baseline OOM boundary. Nevertheless, it is not yet an optimized attention engine and the present implementation should be reckon as a reference recovery system. Below we specify three limitations and each marks a direction we regard as open rather than a defect of the mechanism.

\emph{Kernel coverage.} The native kernel covers the head dimensions FlashAttention-2 does in the forward pass from $32$ to $256$. But the backward currently supports only $64$ and $128$. Expanding that range is necessary for a production version. Two further directions follow: a faster kernel with better memory scaling and approximated kernels admitted to the interface. In addition, converting manually is labor-intensive, and automating it with a tool or an agent that rewrites an attention implementation into a compatible one is a promising direction.

\emph{Scheduling.} The evaluation in this work uses the simplest single-device uniform schedule, which leaves plenty of room for optimization. The subproblems are independent by construction, so a scheduler may execute several concurrently, overlap transfer with computation, and place them across multiple or heterogeneous devices (Section~\ref{sec:workload_sched}). A distributed implementation is the natural next step and we regard this as an important engineering and system design problem.

\emph{System integration.} The experiments isolate attention, while real OOMs can arise from activation caches, temporary buffers, optimizer state, and other model components. Stream-CQSA relocates some memory pressure rather than removing it, so end-to-end integration in a language model is needed to decide where recovery should trigger. Edge devices are another relevant setting, where fixed and modest memory budgets may make a time-for-memory tradeoff worthwhile.

Repeated decomposition can make each in-flight subproblem as small as the budget permits, aided by host-resident fp32 accumulators and a cap on resident subproblems. What can not be shrunk is the full-length data itself such as $Q$, $K$ and $V$ and the final attention output $O$. Stream-CQSA relocates those tensors into host memory, which is larger and cheaper but still finite. Therefore, extremely long sequences can exhaust host memory or produce outputs that fill the GPU. This intrinsic limit is outside what this mechanism can reach.


\clearpage
\bibliographystyle{unsrtnat}
\bibliography{references}

@article{vaswani2017attention,
  title={Attention is all you need},
  author={Vaswani, Ashish and Shazeer, Noam and Parmar, Niki and Uszkoreit, Jakob and Jones, Llion and Gomez, Aidan N and Kaiser, {\L}ukasz and Polosukhin, Illia},
  journal={Advances in neural information processing systems},
  volume={30},
  year={2017}
}

@article{lam1991search,
  title={The search for a finite projective plane of order 10},
  author={Lam, Clement WH},
  journal={The American mathematical monthly},
  volume={98},
  number={4},
  pages={305--318},
  year={1991},
  publisher={Taylor \& Francis}
}

@inproceedings{bian2021efficient,
  title={An Efficient Systematic Approach to Find All Cyclic Quorum Sets with All-pairs Property},
  author={Bian, Yiming and Somani, Arun K},
  booktitle={2021 IEEE International Conference on Big Data (Big Data)},
  pages={197--206},
  year={2021},
  organization={IEEE}
}

@article{kaplan2020scaling,
  title={Scaling laws for neural language models},
  author={Kaplan, Jared and McCandlish, Sam and Henighan, Tom and Brown, Tom B and Chess, Benjamin and Child, Rewon and Gray, Scott and Radford, Alec and Wu, Jeffrey and Amodei, Dario},
  journal={arXiv preprint arXiv:2001.08361},
  year={2020}
}

@article{beltagy2020longformer,
  title={Longformer: The long-document transformer},
  author={Beltagy, Iz and Peters, Matthew E and Cohan, Arman},
  journal={arXiv preprint arXiv:2004.05150},
  year={2020}
}

@article{kitaev2020reformer,
  title={Reformer: The efficient transformer},
  author={Kitaev, Nikita and Kaiser, {\L}ukasz and Levskaya, Anselm},
  journal={arXiv preprint arXiv:2001.04451},
  year={2020}
}

@article{wang2020linformer,
  title={Linformer: Self-attention with linear complexity},
  author={Wang, Sinong and Li, Belinda Z and Khabsa, Madian and Fang, Han and Ma, Hao},
  journal={arXiv preprint arXiv:2006.04768},
  year={2020}
}

@misc{claude,
  title={The Claude 3 Model Family: Opus, Sonnet, Haiku},
  author={},
  howpublished = {\url{https://www-cdn.anthropic.com/de8ba9b01c9ab7cbabf5c33b80b7bbc618857627/Model_Card_Claude_3.pdf}}
}

@article{comanici2025gemini,
  title={Gemini 2.5: Pushing the frontier with advanced reasoning, multimodality, long context, and next generation agentic capabilities},
  author={Comanici, Gheorghe and Bieber, Eric and Schaekermann, Mike and Pasupat, Ice and Sachdeva, Noveen and Dhillon, Inderjit and Blistein, Marcel and Ram, Ori and Zhang, Dan and Rosen, Evan and others},
  journal={arXiv preprint arXiv:2507.06261},
  year={2025}
}

@article{guo2025deepseek,
  title={Deepseek-r1: Incentivizing reasoning capability in llms via reinforcement learning},
  author={Guo, Daya and Yang, Dejian and Zhang, Haowei and Song, Junxiao and Wang, Peiyi and Zhu, Qihao and Xu, Runxin and Zhang, Ruoyu and Ma, Shirong and Bi, Xiao and others},
  journal={arXiv preprint arXiv:2501.12948},
  year={2025}
}

@article{yang2025qwen3,
  title={Qwen3 technical report},
  author={Yang, An and Li, Anfeng and Yang, Baosong and Zhang, Beichen and Hui, Binyuan and Zheng, Bo and Yu, Bowen and Gao, Chang and Huang, Chengen and Lv, Chenxu and others},
  journal={arXiv preprint arXiv:2505.09388},
  year={2025}
}

@article{dao2022flashattention,
  title={Flashattention: Fast and memory-efficient exact attention with io-awareness},
  author={Dao, Tri and Fu, Dan and Ermon, Stefano and Rudra, Atri and R{\'e}, Christopher},
  journal={Advances in neural information processing systems},
  volume={35},
  pages={16344--16359},
  year={2022}
}

@article{dao2023flashattention,
  title={Flashattention-2: Faster attention with better parallelism and work partitioning},
  author={Dao, Tri},
  journal={arXiv preprint arXiv:2307.08691},
  year={2023}
}

@article{xiao2023efficient,
  title={Efficient streaming language models with attention sinks},
  author={Xiao, Guangxuan and Tian, Yuandong and Chen, Beidi and Han, Song and Lewis, Mike},
  journal={arXiv preprint arXiv:2309.17453},
  year={2023}
}

@article{ding2023longnet,
  title={Longnet: Scaling transformers to 1,000,000,000 tokens},
  author={Ding, Jiayu and Ma, Shuming and Dong, Li and Zhang, Xingxing and Huang, Shaohan and Wang, Wenhui and Zheng, Nanning and Wei, Furu},
  journal={arXiv preprint arXiv:2307.02486},
  year={2023}
}

@article{zaheer2020big,
  title={Big bird: Transformers for longer sequences},
  author={Zaheer, Manzil and Guruganesh, Guru and Dubey, Kumar Avinava and Ainslie, Joshua and Alberti, Chris and Ontanon, Santiago and Pham, Philip and Ravula, Anirudh and Wang, Qifan and Yang, Li and others},
  journal={Advances in neural information processing systems},
  volume={33},
  pages={17283--17297},
  year={2020}
}

@inproceedings{katharopoulos2020transformers,
  title={Transformers are rnns: Fast autoregressive transformers with linear attention},
  author={Katharopoulos, Angelos and Vyas, Apoorv and Pappas, Nikolaos and Fleuret, Fran{\c{c}}ois},
  booktitle={International conference on machine learning},
  pages={5156--5165},
  year={2020},
  organization={PMLR}
}

@article{choromanski2020rethinking,
  title={Rethinking attention with performers},
  author={Choromanski, Krzysztof and Likhosherstov, Valerii and Dohan, David and Song, Xingyou and Gane, Andreea and Sarlos, Tamas and Hawkins, Peter and Davis, Jared and Mohiuddin, Afroz and Kaiser, Lukasz and others},
  journal={arXiv preprint arXiv:2009.14794},
  year={2020}
}

@article{chen2023longlora,
  title={Longlora: Efficient fine-tuning of long-context large language models},
  author={Chen, Yukang and Qian, Shengju and Tang, Haotian and Lai, Xin and Liu, Zhijian and Han, Song and Jia, Jiaya},
  journal={arXiv preprint arXiv:2309.12307},
  year={2023}
}

@inproceedings{qiu2020blockwise,
  title={Blockwise self-attention for long document understanding},
  author={Qiu, Jiezhong and Ma, Hao and Levy, Omer and Yih, Wen-tau and Wang, Sinong and Tang, Jie},
  booktitle={Findings of the Association for Computational Linguistics: EMNLP 2020},
  pages={2555--2565},
  year={2020}
}

@article{bian2025cqs,
  title={CQS-Attention: Scaling Up the Standard Attention Computation for Infinitely Long Sequences},
  author={Bian, Yiming and Somani, Arun K},
  journal={IEEE Access},
  year={2025},
  publisher={IEEE}
}

@book{van2001course,
  title={A course in combinatorics},
  author={Van Lint, Jacobus Hendricus and Wilson, Richard Michael},
  year={2001},
  publisher={Cambridge university press}
}

@article{singer1938theorem,
  title={A theorem in finite projective geometry and some applications to number theory},
  author={Singer, James},
  journal={Transactions of the American Mathematical Society},
  volume={43},
  number={3},
  pages={377--385},
  year={1938},
  publisher={JSTOR}
}

@article{macwilliams1973existence,
  title={On the existence of a projective plane of order 10},
  author={MacWilliams, F Jessie and Sloane, Neil JA and Thompson, John G},
  journal={Journal of Combinatorial Theory, Series A},
  volume={14},
  number={1},
  pages={66--78},
  year={1973},
  publisher={Elsevier}
}

@article{stinson2008combinatorial,
  title={Combinatorial designs: constructions and analysis},
  author={Stinson, Douglas R},
  journal={ACM SIGACT News},
  volume={39},
  number={4},
  pages={17--21},
  year={2008},
  publisher={ACM New York, NY, USA}
}

@article{rabe2021self,
  title={Self-attention does not need $ O (n^2) $ memory},
  author={Rabe, Markus N and Staats, Charles},
  journal={arXiv preprint arXiv:2112.05682},
  year={2021}
}

@inproceedings{li2023sequence,
  title={Sequence parallelism: Long sequence training from system perspective},
  author={Li, Shenggui and Xue, Fuzhao and Baranwal, Chaitanya and Li, Yongbin and You, Yang},
  booktitle={Proceedings of the 61st Annual Meeting of the Association for Computational Linguistics (Volume 1: Long Papers)},
  pages={2391--2404},
  year={2023}
}

@inproceedings{liu2024ringattention,
  title={Ringattention with blockwise transformers for near-infinite context},
  author={Liu, Hao and Zaharia, Matei and Abbeel, Pieter},
  booktitle={International Conference on Learning Representations},
  volume={2024},
  pages={3992--4008},
  year={2024}
}

@article{brandon2023striped,
  title={Striped attention: Faster ring attention for causal transformers},
  author={Brandon, William and Nrusimha, Aniruddha and Qian, Kevin and Ankner, Zachary and Jin, Tian and Song, Zhiye and Ragan-Kelley, Jonathan},
  journal={arXiv preprint arXiv:2311.09431},
  year={2023}
}

@article{jacobs2023deepspeed,
  title={Deepspeed ulysses: System optimizations for enabling training of extreme long sequence transformer models},
  author={Jacobs, Sam Ade and Tanaka, Masahiro and Zhang, Chengming and Zhang, Minjia and Song, Shuaiwen Leon and Rajbhandari, Samyam and He, Yuxiong},
  journal={arXiv preprint arXiv:2309.14509},
  year={2023}
}

@article{li2023distflashattn,
  title={Distflashattn: Distributed memory-efficient attention for long-context llms training},
  author={Li, Dacheng and Shao, Rulin and Xie, Anze and Xing, Eric P and Ma, Xuezhe and Stoica, Ion and Gonzalez, Joseph E and Zhang, Hao},
  journal={arXiv preprint arXiv:2310.03294},
  year={2023}
}

@article{milakov2018online,
  title={Online normalizer calculation for softmax},
  author={Milakov, Maxim and Gimelshein, Natalia},
  journal={arXiv preprint arXiv:1805.02867},
  year={2018}
}

\appendix
\clearpage
\section{Native kernel implementation}
\label{sec:native_kernel}

The preceding sections specify the decomposition, exact recomposition, and depth-selection policy. They do not by themselves yield a competitive kernel. A reference implementation that materializes the local score matrix (Appendix~\ref{sec:appendix_experiments}) demonstrates correctness but runs roughly an order of magnitude slower than a production attention path.

The native kernel addresses three sources of slowdown. First, recomposition through unnormalized sums can overflow fp32 because it requires exponentiating log-sum-exp values. Second, every subproblem must apply the CQS mask cheaply enough that bookkeeping does not dominate. Third, decomposition multiplies kernel launches and host--device transfers, so the schedule must avoid waiting on the host once per subproblem.

\subsection{Overflow-free recomposition}
\label{sec:kernel_overflow}

The direct recomposition uses an unnormalized numerator and denominator. For subproblem $i$ with local scores $R_i$,
\begin{equation} \mathrm{Num}_i = \exp(R_i)V_i \qquad \mathrm{Den}_i = \mathrm{row\_sum}\,\exp(R_i) \qquad O = \Big(\textstyle\sum_i \mathrm{Num}_i\Big)\Big/\Big(\textstyle\sum_i \mathrm{Den}_i\Big)
\label{eq:numden}
\end{equation}
This is natural because many local kernels return normalized output $\texttt{out}_i$ and log-sum-exp $\texttt{lse}_i$, from which one can reconstruct $\mathrm{Den}_i=\exp(\texttt{lse}_i)$ and $\mathrm{Num}_i=\texttt{out}_i\odot\mathrm{Den}_i$. Equation~\eqref{eq:numden} is algebraically correct but numerically unsafe. The exponential overflows fp32 when $\texttt{lse}>\ln(\textsc{flt\_max})\approx88.72$. Beyond that point $\mathrm{Den}_i$ becomes $\infty$, the ratio becomes $\infty/\infty$, and the result can silently become zero or NaN. The backward substitution has the same problem through $\mathbf{d}\mathrm{Num}=\mathbf{d}O/\mathrm{Den}$ and $\mathbf{d}\mathrm{Den}=-\mathrm{row\_sum}(\mathbf{d}O\odot\mathrm{Num})/\mathrm{Den}^2$. We avoid overflow in both passes.

\paragraph{Forward max-shifted merge.} Each subproblem returns three statistics rather than a normalized output, namely a running row maximum $m$, a running sum $\ell$ of $\exp(\text{score}-m)$, and a running weighted sum $\mathrm{acc}$ of $\exp(\text{score}-m)V$. Merging a contribution $(\mathrm{acc}_i,\ell_i,m_i)$ re-bases both sides onto $m'=\max(m,m_i)$. Thus,
\begin{equation}
\mathrm{acc} \leftarrow \mathrm{acc}\,e^{m-m'} + \mathrm{acc}_i\,e^{m_i-m'} 
\qquad
\ell \leftarrow \ell\,e^{m-m'} + \ell_i\,e^{m_i-m'}
\qquad m \leftarrow m'
\label{eq:merge}
\end{equation}
Since $m'$ is the larger maximum, both exponents are at most zero, so no intermediate overflows. The final output $O=\mathrm{acc}/\ell$ and global log-sum-exp $\texttt{lse}=m+\log\ell$ are formed only after all subproblems are merged. No subproblem computes $\exp(\texttt{lse})$. The merge is order-independent, allowing subproblems to finish in scheduler order. Eq.~\eqref{eq:merge} is the standard online softmax recurrence~\cite{milakov2018online}. The important point is that Stream-CQSA must carry the same three statistics across subproblems. A framework that collects only $\texttt{out}_i$ and $\texttt{lse}_i$ would still need the unsafe exponential in Eq.~\eqref{eq:numden}.

\paragraph{Backward global-lse reformulation.} Substituting $\mathrm{Num}=\mathrm{Den}\odot O$ and folding $\mathrm{Den}$ into $P$ gives
\begin{equation}
P[q,k] = \exp\!\big(s[q,k] - \texttt{lse}_{\text{global}}[q]\big) \le 1 \qquad \text{for all } q,k.
\label{eq:globallse}
\end{equation}
The forward already produces $\texttt{lse}_{\text{global}}$, so keeping it is free. Each subproblem receives this value and runs the standard softmax backward on the pairs it retains. Because CQS covers each query--key pair exactly once, subproblem gradients add directly. Also, $P\le1$ by construction, so intermediates cannot overflow regardless of score range. Eq.~\eqref{eq:globallse} is algebraically identical to Algorithm~\ref{algo:bwd}. We state the implemented form separately because a literal implementation of the algorithm would reintroduce the overflow above.

\paragraph{Statistics in fp32.} The triple $(\mathrm{acc},\ell,m)$ and the merge in Eq.~\eqref{eq:merge} are computed in fp32 for all input dtypes, and the accumulator is written in fp32 instead of being rounded to 16 bits. This prevents merge error from compounding (Section~\ref{sec:exp_accuracy}) but makes the accumulator larger than $Q$, $K$, and $V$ combined at 16-bit input precision. Its size also does not shrink with decomposition depth, driving the memory behavior in Section~\ref{sec:exp_recovery}. One corner case remains. A subsequence may retain no keys for some query row. The sentinel $m=-\infty$ would make Eq.~\eqref{eq:merge} compute $\exp(-\infty-(-\infty))$, producing NaN, so non-finite contributions are replaced by null contributions before merge.

\subsection{Making the CQS mask cheap}
\label{sec:kernel_mask}

A subsequence gathers whole chunks, but not every local query--key pair belongs to that subproblem. The CQS mask (Section~\ref{sec:cqs_mask}) excludes the others. Each token carries a small bit vector recording its CQS groups, and a local pair $(q,k)$ is retained exactly when $\texttt{bits}[q]\wedge\texttt{bits}[k]=0$. The generic implementation tests this predicate per element and reads both bit vectors from global memory in the innermost loop. With an all-zero mask, which excludes nothing and performs no useful masking, this path made the forward $4.3\times$ slower and backward $9.1\times$ slower than the unmodified kernel. Three techniques recover that overhead.

\paragraph{Block summaries and an $O(1)$ tile verdict.} An arbitrary mask requires pairwise tests, but the CQS mask is chunk-structured. Since each subsequence is a union of contiguous chunks, neighboring tokens share bits except at chunk boundaries. We exploit this with a summary test that classifies a whole score-matrix tile in constant time from precomputed block summaries. Here a \emph{block} is a one-dimensional run of $64$ tokens, distinct from a two-dimensional attention tile. For each block we store the bitwise OR and AND of its token group bits. The OR records bits present in any token, and the AND records bits present in every token. Summaries depend only on the decomposition shape and are computed once per subproblem with the mask.

A FlashAttention \emph{tile} is a $128\times128$ score region. Its rows span two token blocks and its columns span two more, so the summary test reads four block summaries. OR-ing the row-block ORs gives $\mathrm{OR}(\text{rows})$, while AND-ing the row-block ANDs gives $\mathrm{AND}(\text{rows})$. Columns are combined similarly. Two bitwise conjunctions then classify the tile before any per-element work runs.

A tile is \emph{clear} when $\mathrm{OR}(\text{rows})\wedge \mathrm{OR}(\text{cols})=0$. No row bit appears in any column, so no pair is masked and the tile uses the unmodified FlashAttention path. It is \emph{fully masked} when $\mathrm{AND}(\text{rows})\wedge\mathrm{AND}(\text{cols})\ne 0$. A shared bit appears in every row and column, so every pair is excluded and the tile contributes zero. These tests cannot create an incorrect keep/drop decision. Their only possible failure is conservative. A tile may be classified as mixed and sent to the per-element path even though a finer test could have resolved it.

\begin{figure}[t]
\centering
\includegraphics[width=\linewidth]{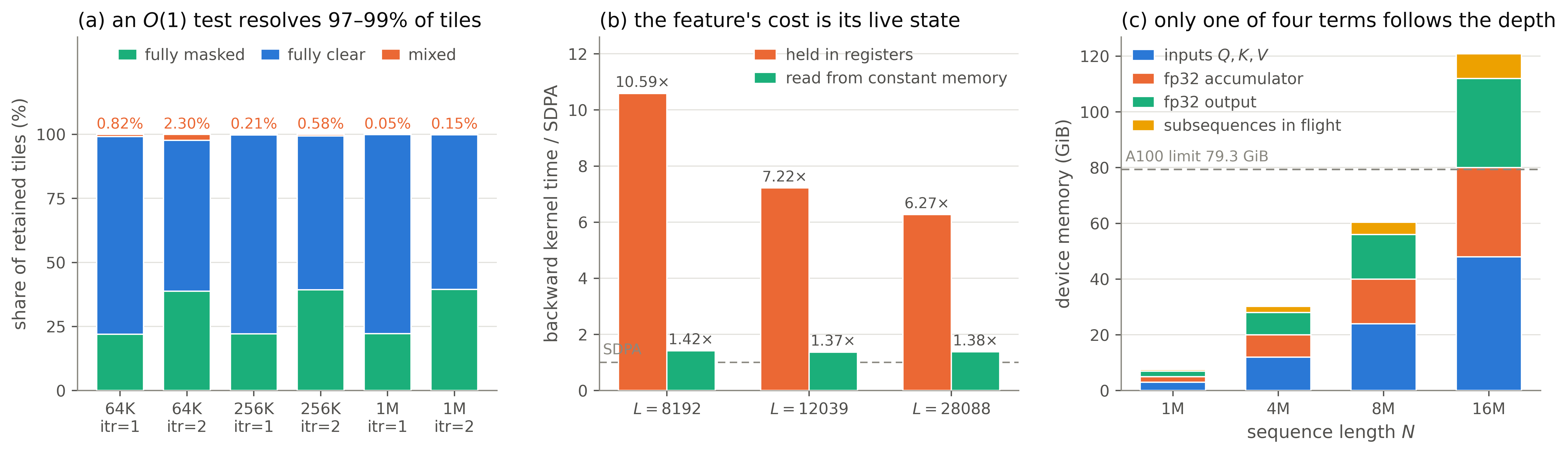}
\caption{The three implementation decisions measured.
\textbf{(a)} Census of the $128\times128$ tiles that causal masking retains.
Because the CQS mask is block-structured, the $O(1)$ summary test resolves
$97$--$99\%$ of them, and the per-element path runs on the small remainder
printed above each bar. That remainder falls as $N$ grows, since a tile can be
mixed only where it straddles a chunk boundary.
\textbf{(b)} Backward kernel time relative to SDPA at identical shape, measured
with an all-zero CQS mask so that what is isolated is the cost of carrying the
feature rather than of any masking it performs. Holding the mask state as object
members pushed a register-saturated kernel into local-memory spill. Reading the
same state from constant memory restores near parity, with the masking logic
unchanged.
\textbf{(c)} The four $O(N)$ contributions to device memory. Only the in-flight
subsequences shrink as $\texttt{itr}$ grows, so raising the depth alone cannot
avert an out-of-memory failure once the other three dominate, which is the
observation that governs the recovery behavior in
Section~\ref{sec:exp_recovery}.}
\label{fig:kernel_panels}
\end{figure}

In practice, almost every tile is resolved by the summary test. Figure~\ref{fig:kernel_panels}a reports a pooled census at $N=65\,536$ over causal-retained $128\times128$ tiles. At $\texttt{itr}=1$, the seven subproblems yield $22.0\%$ fully masked, $77.2\%$ clear, and $0.82\%$ mixed. At $\texttt{itr}=2$, the forty-nine subproblems yield $38.8\%$ fully masked, $58.9\%$ clear, and $2.30\%$ mixed. Thus $97$--$99\%$ of tiles never reach the per-element path. The masked/clear split varies by subproblem, especially at $\texttt{itr}=2$, but the mixed fraction is small and decreases with sequence length.

This follows from counting. Let $n_b=\lceil L/128\rceil$ be the number of tiles along one side of a local score matrix, so the matrix has $n_b^2$ tiles. Tokens within a chunk share bits, so a tile can be mixed only if it straddles a chunk boundary. A subsequence with $l$ chunks has at most $l$ such boundaries, and each boundary affects at most one row and one column of tiles. Mixed tiles are therefore $O(l\,n_b)$ out of $O(n_b^2)$, giving a fraction on the order of $l/n_b$ that vanishes with sequence length.

Measurements match this scaling and stay below the bound. At $\texttt{itr}=2$, the mixed fractions are $2.30\%$ at $N=65\,536$ ($n_b=95$), $0.58\%$ at $N=262\,144$ ($n_b=377$), and $0.15\%$ at $N=1\,048\,576$ ($n_b=1505$), versus predicted bounds of $3.16\%$, $0.80\%$, and $0.20\%$. The bound overshoots because it charges every boundary a full tile row and column, while boundaries on tile edges cost nothing. The scaling is what matters. At $N=1\,048\,576$, only about one tile in $670$ takes the general path, so optimizing that path further could recover at most $0.15\%$ of runtime.

\paragraph{Register budget.} The largest single effect was register pressure. Each CUDA thread has a compile-time register allocation capped at $255$ registers. If more values are live than fit, the compiler spills to CUDA local memory, which resides in off-chip DRAM. A spill inside an innermost loop turns a register read into repeated memory traffic.

FlashAttention kernels already use nearly all available registers to keep score tiles, softmax statistics, accumulators, and input fragments out of memory. Our backward kernel compiled at $254$--$255$ registers per thread. The mask added nine values, including pointers that cost two registers each. Storing them as members of a mask object kept them live for the whole kernel and caused spills. We moved the same values into the kernel parameter block, which resides in constant memory, and read them only at use sites. The masking logic did not change. At $L=8\,192$, backward time fell from $10.59\times$ SDPA to $1.42\times$ (Figure~\ref{fig:kernel_panels}b). At $L=28\,088$, it fell from $6.27\times$ to $1.38\times$. In a register-saturated kernel, the cost of a feature can be dominated by live state rather than arithmetic.

\paragraph{Skipping fully-masked tiles.} A fully masked tile contributes zero, making its five GEMMs redundant. In the backward pass, however, the $m$-block loop also performs stateful work that must still run. It accumulates and writes back $\texttt{dq\_accum}$, swaps shared-memory double buffers, issues \texttt{cp.async} prefetches, and reaches synchronization barriers. We therefore guard the GEMMs, mask application, and exponential individually while leaving stateful steps active. This is safe because the tile verdict depends only on block indices and summaries, not thread indices, so every thread in the block takes the same branch and reaches the same barriers.

Correctness does not require every downstream guard to fire. Accumulators are cleared at the loop head. If the first GEMM is skipped, the score block $S$ remains zero. If the exponential is skipped, the softmax weights $P$ remain zero. The score gradient is $\mathbf{d}S=P\odot(\mathbf{d}P-D)$, so $\mathbf{d}S=0$ regardless of $\mathbf{d}P$ or the per-row correction $D$. Any remaining GEMMs that use $\mathbf{d}S$ then add zero. This explains why, measured in isolation, a real CQS mask is $1.42\times$ faster than an all-zero mask at $N=262\,144$ and $\texttt{itr}=2$. The all-zero mask excludes nothing, so no tile is fully masked and no arithmetic can be skipped.

\subsection{Scheduling, occupancy and residency}
\label{sec:kernel_sched}

Subproblems are independent and run on separate CUDA streams. The useful number in flight is limited by memory and device saturation. The kernel assigns one thread block to each $128$-query strip for every head and batch element, so a subsequence of length $L$ issues about $\lceil L/128\rceil\cdot B\cdot H$ thread blocks. Once that count exceeds the number of streaming multiprocessors, another stream cannot add arithmetic throughput, though it can still overlap gather and transfer with compute. At the relevant sizes the device is heavily oversubscribed. An A100 has $108$ SMs, while one subproblem at $N=32\,768$ and $H=8$ issues $880$ blocks. The scheduler therefore caps concurrency at two streams whenever one task alone issues at least twice the SM count, and allows more only for smaller tasks. We also gather chunks in ascending order so local token order matches global order. The unmodified FlashAttention causal mask is then correct and composes with the CQS mask by conjunction.

In this kernel, $\texttt{n\_cap}$ is chosen for memory rather than throughput because one subproblem already saturates the device. Calibration admits as many resident subsequences as the budget allows, which is seven or eight at shorter measured lengths and one from $N=2\,097\,152$ onward. In the recovery regime, $\texttt{n\_cap}=1$ and the guardrail reduces to the depth rule. The branch that decreases depth requires all subproblems to be resident with spare memory. That condition never occurs in our measurements, so reported depth is always the first successful depth.

Depth can be explicit or automatic. An explicit depth bypasses the guardrail. If it OOMs, the failure is reported. This is how fixed-depth rows in Section~\ref{sec:results} are measured, so a row labeled $\texttt{itr}=1$ reports that depth even where it does not fit. Automatic mode begins at $\texttt{itr}=1$ and applies the guardrail, reaching the smallest depth at which no subproblem exhausts memory.

The other axis is residency, meaning where tensors are kept. Residency does not change the computation, only device memory and therefore maximum sequence length. Four $O(N)$ quantities can occupy device memory (Figure~\ref{fig:kernel_panels}c), including inputs $Q,K,V$, the fp32 accumulator, the returned output, and in-flight subsequences. Only in-flight subsequences shrink with depth. The other terms shrink only when moved to host memory. These terms set a device-memory floor that no decomposition depth can cross.

The device-resident boundary depends on memory budget, not a fixed $N$. At $B=1$, $H=8$, and $D=64$, forward--backward at $\texttt{itr}=1$ uses about $13.6$ GiB per million tokens, placing the boundary near $5.8\mathrm{M}$ on the $79.3$ GiB available here. This matches the observed pattern. $4\text{M}$ completes and $8\text{M}$ does not. The same estimate gives roughly $2.9\mathrm{M}$ on a $40$ GiB card and $10\text{M}$ on a $140$ GiB card. Forward--backward sets the boundary because a training step needs both passes.

Below that boundary, the device-resident configuration is faster. Above it, host residency becomes necessary, and its time cost is already small. Moving the accumulator to host reduces peak device memory by $2.50\times$ in the forward pass and $1.79\times$ in forward--backward at every measured length. Its runtime penalty shrinks with $N$, from $1.49\times$ at $N=1\mathrm{M}$ to $1.05\times$ at $N=8\mathrm{M}$ forward, and from $1.17\times$ to $1.04\times$ at $N=4\mathrm{M}$ forward--backward. Transfer is linear while attention is quadratic, so the penalty is largest where a monolithic call would already work and only a few percent in the intended regime. We therefore use host-resident accumulators by default.

Forward--backward runs out first because it must accumulate three full-sequence fp32 tensors, one each for $\mathbf{d}Q$, $\mathbf{d}K$, and $\mathbf{d}V$. The forward pass accumulates one fp32 $[B,N,H,D]$ tensor plus row statistics. The backward pass needs three such tensors, or $12$ bytes per element of $[B,N,H,D]$ versus the forward's $4$. Deeper decomposition cannot shrink these full-sequence tensors, so the backward boundary arrives while forward still has room.

\section{Additional reference-kernel experiments}
\label{sec:appendix_experiments}

This appendix tests whether recovery is a framework property rather than an artifact of the native kernel. We convert two reference PyTorch kernels to the Stream-CQSA interface, an exact dense kernel and an approximate LongNet-style kernel~\cite{ding2023longnet}. We write the unwrapped kernels as \emph{dense} and \emph{LongNet}, and the wrapped paths as $\mathcal{S}(\text{dense})$ and $\mathcal{S}(\text{LongNet})$. These implementations are not optimized. They establish compatibility requirements and recovery behavior. Each local subproblem must accept the combined causal and CQS mask, and the inner kernel must expose local numerator/denominator statistics for recomposition. They intentionally follow the algorithms directly, without the overflow-avoidance mechanisms of Section~\ref{sec:kernel_overflow}, so the reported speed, memory, and numerical behavior should be read as properties of these reference paths rather than best-performance claims.

Unless otherwise stated, these experiments use causal attention with $B=1$, $H=8$, $D=64$, bf16 precision, and one NVIDIA A100 80 GiB GPU. Synthetic Q/K/V tensors use fixed random seeds, are drawn from a zero-mean Gaussian scaled by $0.05$, and are cast to the experiment precision. All Stream-CQSA paths use $c=7$, $\mathcal{I}=(0,1,3)$, and $\texttt{itr}=1$.

\paragraph{Output preservation}

Making a kernel Stream-CQSA compatible should not change its attention result. We compare each wrapped path with its unwrapped counterpart for forward outputs and backward gradients, using $N=4096$ and 20 random Q/K/V draws per configuration. Forward errors cover the full output tensor $[B,N,H,D]$. Backward errors are computed jointly over $\mathbf{d}Q$, $\mathbf{d}K$, and $\mathbf{d}V$.

Table~\ref{tab:output_preservation} shows that both reference kernels are preserved within numerical precision after decomposition and recomposition. Table~\ref{tab:bwd_gradient_decomposition} breaks backward errors down by gradient tensor. The largest low-precision maximum errors come from $\mathbf{d}V$, especially in bf16, because $\mathbf{d}V=P^\top\mathbf{d}O$ accumulates unit-scale upstream gradients over many query positions. fp16 is more vulnerable to overflow in the direct exponential, while bf16 has wider exponent range but coarser mantissa precision. Mean absolute errors remain small ($\approx 10^{-5}$ for bf16 and $\approx 10^{-6}$ for fp16), and fp32 errors are near numerical noise. These runs deliberately omit the max-shifted merge of Section~\ref{sec:kernel_overflow}. The visible low-precision errors motivate that implementation requirement.

\begin{table}[htbp]
\centering
\caption{Forward-output and backward-gradient preservation across precision
settings.}
\label{tab:output_preservation}
\begin{tabular}{ccccc}
\toprule
Pair & Pass & Precision
& Max abs. error & Mean abs. error \\
\midrule
\multirow{6}{*}{\shortstack{dense\\vs\\$\mathcal{S}(\text{dense})$}}
& \multirow{3}{*}{fwd}
& fp32 & $(1.56 \pm 0.20)\times 10^{-8}$
& $(7.99 \pm 0.16)\times 10^{-10}$ \\
& & bf16 & $(1.56 \pm 0.19)\times 10^{-8}$
& $(7.98 \pm 0.14)\times 10^{-10}$ \\
& & fp16 & $(1.51 \pm 0.10)\times 10^{-8}$
& $(7.99 \pm 0.13)\times 10^{-10}$ \\
\cmidrule(lr){2-5}
& \multirow{3}{*}{bwd}
& fp32 & $(9.26 \pm 1.60)\times 10^{-6}$
& $(6.11 \pm 0.08)\times 10^{-9}$ \\
& & bf16 & $(1.88 \pm 0.64)\times 10^{-2}$
& $(1.08 \pm 0.01)\times 10^{-5}$ \\
& & fp16 & $(2.25 \pm 0.72)\times 10^{-3}$
& $(1.35 \pm 0.02)\times 10^{-6}$ \\

\midrule
\multirow{6}{*}{\shortstack{LongNet\\vs\\$\mathcal{S}(\text{LongNet})$}}
& \multirow{3}{*}{fwd}
& fp32 & $(1.17 \pm 0.12)\times 10^{-8}$
& $(5.66 \pm 0.04)\times 10^{-10}$ \\
& & bf16 & $(1.18 \pm 0.15)\times 10^{-8}$
& $(5.66 \pm 0.04)\times 10^{-10}$ \\
& & fp16 & $(1.14 \pm 0.08)\times 10^{-8}$
& $(5.67 \pm 0.03)\times 10^{-10}$ \\
\cmidrule(lr){2-5}
& \multirow{3}{*}{bwd}
& fp32 & $(3.23 \pm 0.69)\times 10^{-6}$
& $(4.25 \pm 0.03)\times 10^{-9}$ \\
& & bf16 & $(1.72 \pm 0.48)\times 10^{-2}$
& $(1.86 \pm 0.01)\times 10^{-5}$ \\
& & fp16 & $(2.25 \pm 0.72)\times 10^{-3}$
& $(2.33 \pm 0.02)\times 10^{-6}$ \\
\bottomrule
\end{tabular}
\end{table}

\begin{table}[htbp]
\centering
\caption{Backward-gradient preservation decomposed by gradient tensor. Errors
are computed after converting both tensors to fp32.}
\label{tab:bwd_gradient_decomposition}
\resizebox{\textwidth}{!}{%
\begin{tabular}{ccccc}
\toprule
Pair & Precision
& $\mathbf{d}Q$ max abs. error
& $\mathbf{d}K$ max abs. error
& $\mathbf{d}V$ max abs. error \\
\midrule
\multirow{3}{*}{\shortstack{dense\\vs\\$\mathcal{S}(\mathrm{dense})$}} 
& fp32 & $(2.43 \pm 0.63)\times 10^{-9}$ & $(1.70 \pm 0.46)\times 10^{-8}$ & $(9.26 \pm 1.60)\times 10^{-6}$ \\
& bf16 & $(3.15 \pm 2.08)\times 10^{-6}$ & $(3.36 \pm 0.94)\times 10^{-5}$ & $(1.88 \pm 0.64)\times 10^{-2}$ \\
& fp16 & $(7.87 \pm 5.36)\times 10^{-7}$ & $(3.91 \pm 0.97)\times 10^{-6}$ & $(2.25 \pm 0.72)\times 10^{-3}$ \\
\midrule
\multirow{3}{*}{\shortstack{LongNet\\vs\\$\mathcal{S}(\mathrm{LongNet})$}}
& fp32 & $(1.30 \pm 0.40)\times 10^{-9}$ & $(5.41 \pm 1.31)\times 10^{-9}$ & $(3.23 \pm 0.69)\times 10^{-6}$ \\
& bf16 & $(4.01 \pm 0.85)\times 10^{-6}$ & $(3.20 \pm 0.68)\times 10^{-5}$ & $(1.72 \pm 0.48)\times 10^{-2}$ \\
& fp16 & $(5.25 \pm 1.47)\times 10^{-7}$ & $(4.01 \pm 0.85)\times 10^{-6}$ & $(2.25 \pm 0.72)\times 10^{-3}$ \\
\midrule
\midrule
Pair & Precision
& $\mathbf{d}Q$ mean abs. error
& $\mathbf{d}K$ mean abs. error
& $\mathbf{d}V$ mean abs. error \\
\midrule
\multirow{3}{*}{\shortstack{dense\\vs\\$\mathcal{S}(\mathrm{dense})$}}
& fp32 & $(4.96 \pm 0.03)\times 10^{-11}$ & $(5.48 \pm 0.02)\times 10^{-11}$ & $(1.82 \pm 0.02)\times 10^{-8}$ \\
& bf16 & $(7.56 \pm 0.03)\times 10^{-8}$ & $(7.91 \pm 0.03)\times 10^{-8}$ & $(3.21 \pm 0.03)\times 10^{-5}$ \\
& fp16 & $(1.71 \pm 0.02)\times 10^{-8}$ & $(1.57 \pm 0.03)\times 10^{-8}$ & $(4.01 \pm 0.06)\times 10^{-6}$ \\
\midrule
\multirow{3}{*}{\shortstack{LongNet\\vs\\$\mathcal{S}(\mathrm{LongNet})$}}
& fp32 & $(3.06 \pm 0.003)\times 10^{-11}$ & $(3.31 \pm 0.005)\times 10^{-11}$ & $(1.27 \pm 0.01)\times 10^{-8}$ \\
& bf16 & $(1.46 \pm 0.002)\times 10^{-7}$ & $(1.37 \pm 0.004)\times 10^{-7}$ & $(5.56 \pm 0.02)\times 10^{-5}$ \\
& fp16 & $(2.20 \pm 0.003)\times 10^{-8}$ & $(2.06 \pm 0.005)\times 10^{-8}$ & $(6.94 \pm 0.05)\times 10^{-6}$ \\
\bottomrule
\end{tabular}%
}
\end{table}

\paragraph{OOM boundary and recovery cost}
\label{sec:overhead_breakdown}

We compare each unwrapped kernel with its Stream-CQSA path as sequence length $N$ increases. The wrapped paths should be slower while using less GPU memory. The question is whether they continue after the unwrapped implementation reaches its OOM boundary.

\begin{figure}[htbp]
\centering
\includegraphics[width=\textwidth]{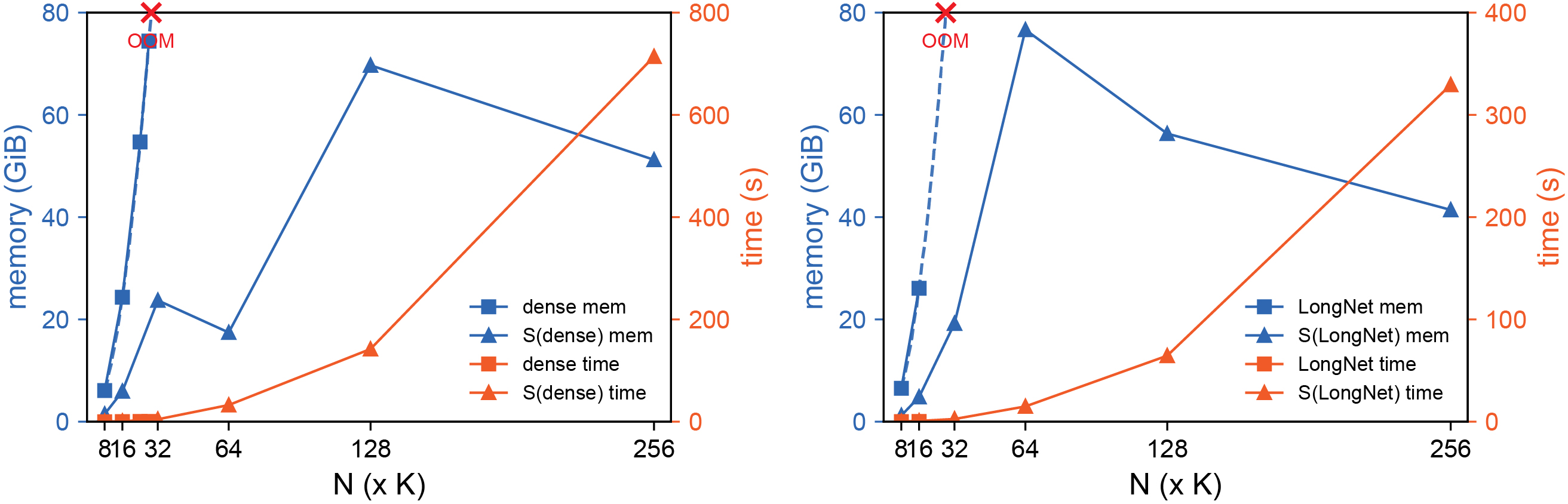}
\caption{Peak allocated GPU memory and wall-clock time for the unwrapped reference kernels and their Stream-CQSA-compatible paths. A memory drop in a Stream-CQSA path indicates an additional decomposition level.}
\label{fig:experiment2}
\end{figure}

Figure~\ref{fig:experiment2} and Table~\ref{tab:exp3_overhead_appendix} compare the dense and LongNet kernels before and after Stream-CQSA wrapping. Peak memory is the increase in PyTorch-allocated GPU memory measured by \texttt{torch.cuda.max\_memory\_allocated()} during attention. This is a reproducible implementation metric, not a theoretical bound or total device-pressure measurement.

The exact dense reference succeeds through $N=28672$ and OOMs at $N=30720$,
whereas $\mathcal{S}(\text{dense})$ completes the tested range through
$N=262144$ by increasing depth. LongNet succeeds through $N=16384$ and OOMs at
$N=32768$, whereas $\mathcal{S}(\text{LongNet})$ also completes through
$N=262144$. The pattern matches the native-kernel results. Wrapping adds
overhead while the original call fits and provides an executable path once it
does not. Because the same recovery behavior appears for native, dense, and
LongNet kernels, it belongs to the decomposition interface rather than a single
implementation. Approximate attention benefits on the same terms as exact
attention. It moves the OOM boundary outward but does not remove it.

\begin{table}[htbp]
\centering
\caption{Recovery-overhead summary for the reference implementations. Peak
allocated memory is the increase in PyTorch-allocated GPU memory during the
measured attention computation. For Stream-CQSA, peak memory and final time
correspond to the final successful decomposition attempt, with full Q/K/V kept
on CPU. Adaptation overhead is total adaptive time minus final successful time.}
\label{tab:exp3_overhead_appendix}
\small
\begin{tabular}{cccccc}
\toprule
Path & $N$ & \texttt{itr} & Peak allocated mem. (GiB) & Final time (s) & Adaptation overhead (s) \\
\midrule
\multirow{5}{*}{dense}
& 8192   & -- & 6.11  & 0.045 & -- \\
& 16384  & -- & 24.34 & 0.179 & -- \\
& 24576  & -- & 54.70 & 0.400 & -- \\
& 28672  & -- & 74.43 & 0.546 & -- \\
& 30720  & -- & OOM   & OOM   & -- \\
\midrule
\multirow{6}{*}{$\mathcal{S}(\text{dense})$}
& 8192   & 1 & 1.49  & 0.507   & 0.110 \\
& 16384  & 1 & 5.95  & 1.043   & 0.112 \\
& 32768  & 1 & 23.74 & 4.766   & 0.113 \\
& 65536  & 2 & 17.45 & 32.597  & 17.105 \\
& 131072 & 2 & 69.70 & 142.132 & 3.883 \\
& 262144 & 3 & 51.23 & 714.132 & 6.764 \\
\midrule
\multirow{6}{*}{LongNet}
& 8192   & -- & 6.55  & 0.077 & -- \\
& 16384  & -- & 26.09 & 0.292 & -- \\
& 32768  & -- & OOM   & OOM   & -- \\
& 65536  & -- & OOM   & OOM   & -- \\
& 131072 & -- & OOM   & OOM   & -- \\
& 262144 & -- & OOM   & OOM   & -- \\
\midrule
\multirow{6}{*}{$\mathcal{S}(\text{LongNet})$}
& 8192   & 1 & 1.22  & 0.331   & 0.103 \\
& 16384  & 1 & 4.84  & 0.780   & 0.113 \\
& 32768  & 1 & 19.22 & 2.680   & 0.137 \\
& 65536  & 1 & 76.66 & 14.978  & 0.105 \\
& 131072 & 2 & 56.35 & 64.565  & 5.397 \\
& 262144 & 3 & 41.43 & 329.482 & 8.519 \\
\bottomrule
\end{tabular}
\end{table}

Both wrapped reference paths keep full Q/K/V tensors in host memory and transfer one current subsequence to the GPU at a time. Local numerator/denominator statistics return to host memory for recomposition in global token coordinates. This conservative policy does not reuse overlapping tensors or pipeline transfers with compute, so it exposes near worst-case overhead for the reference framework implementation.

\paragraph{Forward vs. backward}
\label{sec:appendix_forward_vs_backward}

We also compare forward and backward phases for the two reference Stream-CQSA paths. The forward phase uses \texttt{requires\_grad=True}, so the autograd graph required by backward is constructed. The backward phase is measured as \texttt{loss.backward()} for $\texttt{loss}=(O.\texttt{float}() \odot \textbf{d}O).\texttt{sum()}$. This experiment uses fp16, $B=1$, $H=1$, $D=128$, causal masking, $c=7$,
$\mathcal{I}=(0,1,3)$, and $\texttt{itr}=1$.

\begin{table}[t]
\centering
\caption{Forward and backward profile at $N=65536$ for the reference Stream-CQSA paths.}
\label{tab:appendix_fwd_bwd_raw}
\begin{tabular}{cccc}
\toprule
Path & Pass & Time (s) & Peak mem. (GiB) \\
\midrule
\multirow{2}{*}{$\mathcal{S}(\text{dense})$}
& fwd  & $3.70 \pm 0.11$ & $58.74$ \\
& bwd & $1.00 \pm 0.03$ & $57.24$ \\
\midrule
\multirow{2}{*}{$\mathcal{S}(\text{LongNet})$}
& fwd  & $2.04 \pm 0.37$ & $55.54$ \\
& bwd & $0.69 \pm 0.06$ & $51.93$ \\
\bottomrule
\end{tabular}
\end{table}

Table~\ref{tab:appendix_fwd_bwd_raw} reports $N=65536$, the largest length in this profiling run. In these reference implementations, backward is not the dominant runtime cost. It takes $0.27\times$ the forward time for $\mathcal{S}(\text{dense})$ and $0.34\times$ for $\mathcal{S}(\text{LongNet})$. Peak memory is closer, at $0.97\times$ and $0.94\times$ of forward peak memory, respectively. These ratios are implementation-specific and depend on how the inner kernel saves or recomputes forward state and executes backward.

\section{CQS interest set}
\label{appendix:CQS}

\begin{figure}[h]
  \centering
  \centerline{\includegraphics[width=1\linewidth]{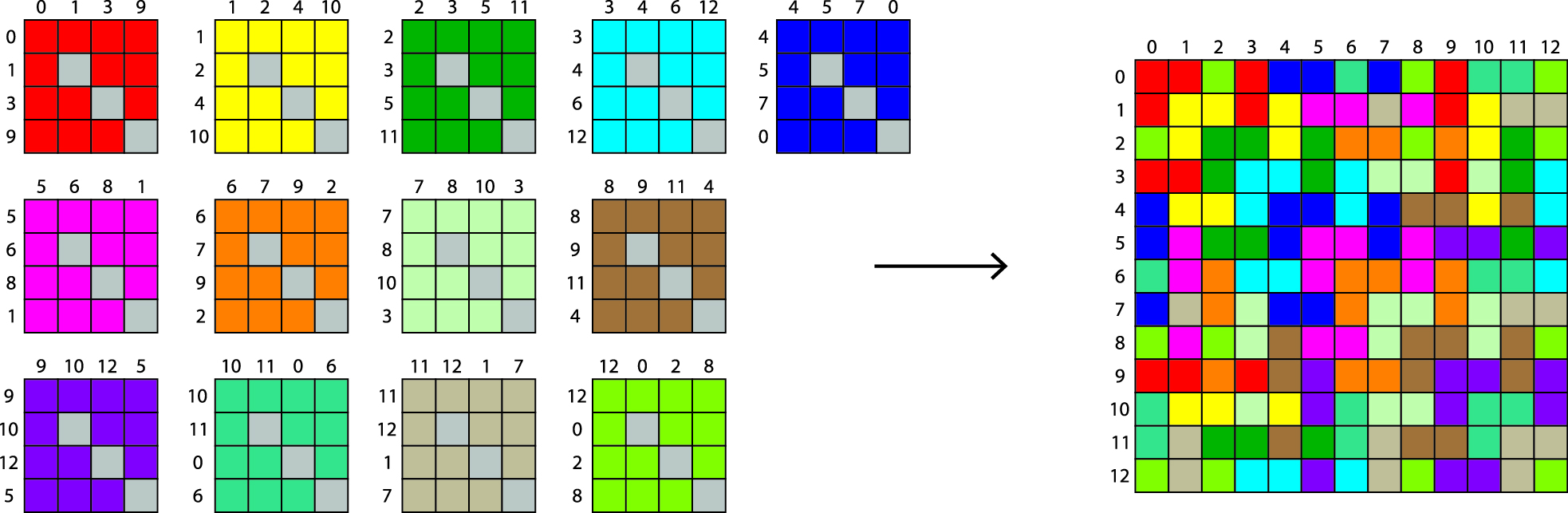}}
  \caption{Full coverage validation for $c=13$ and $\mathcal{I}=(0,1,3,9)$.}
\label{fig:c_13}
\end{figure}

An interest set $\mathcal{I}$ ensures full coverage of pairwise interactions. Table~\ref{tab:interest_set} lists examples for several $l$ values, and Figure~\ref{fig:c_13} validates full coverage for $c=13$ with $\mathcal{I}=(0,1,3,9)$.

Interest sets are not unique and occur in pairs~\cite{bian2021efficient}. Given $(0,1,a_2,a_3,\dots,a_{l-1})$, the paired interest set is $(0,1,c+1-a_{l-1},c+1-a_{l-2},\dots,c+1-a_2)$. For $c=7$, the paired interest set of $(0,1,3)$ is $(0,1,5)$.

Interest sets do not exist for all $l$ values, including $l=7$ and $l=11$ in Table~\ref{tab:interest_set}. Formally, an interest set corresponds to a $(v,k,\lambda)$-\textit{difference set} with $\lambda=1$. This is a $k$-subset $D \subseteq G$ of an abelian group $G$ of order $v$ such that each nonzero $g\in G$ appears exactly $\lambda$ times in the multiset of differences $(x-y \mid x,y\in D)$~\cite{van2001course}. In our notation, $k=l$ and $v=c$.

Let $q=l-1$. Existence of a cyclic $(q^2+q+1,q+1,1)$-\textit{difference set} is equivalent to a projective plane of order $q$ with a cyclic automorphism containing one cycle of length $q+1$~\cite{van2001course}. Such difference sets exist for prime-power $q$ by the Singer construction~\cite{singer1938theorem}. For non-prime-power $q$, existence remains open in many cases and impossible in some.

Lam's problem asks whether a finite projective plane of order $10$ exists. By earlier structural results~\cite{macwilliams1973existence}, this is equivalent to asking whether an interest set of size $l=11$ exists for $c=111$. Lam et al.~\cite{lam1991search} answered no by exhaustive computational search.

\begin{table}[h]
\centering
\caption{Example interest sets for several $l$ values.}
\label{tab:interest_set}
\begin{tabular}{ccccc}
\toprule
$c$   &$l$ & $q$ & prime power &$\mathcal{I}$ \\
\midrule
$7$    & $3$   &$2$ & $2^1$ & $(0,1,3)$           \\
$13$   & $4$   &$3$ & $3^1$ & $(0,1,3,9)$           \\
$21$   & $5$   &$4$ & $2^2$  & $(0,1,4,14,16)$           \\
$31$   & $6$   &$5$ & $5^1$ & $(0,1,3,8,12,18)$           \\
$43$   & $7$   &$6$ & NA & None           \\
$57$   & $8$   &$7$ & $7^1$ & $(0,1,3,13,32,36,43,52)$           \\
$73$   & $9$   &$8$ & $2^3$ & $(0,1,3,7,15,31,36,54, 63)$           \\
$91$   & $10$  &$9$ & $3^2$ & $(0,1,3,9,27,49,56,61,77,81)$           \\
$111$  & $11$  &$10$ & NA & None           \\
$133$  & $12$  &$11$ & $11^1$ & $(0, 1, 3, 12, 20, 34, 38, 81, 88, 94, 104, 109)$ \\
\bottomrule
\end{tabular}%
\end{table}

\section{Chain-rule deductions in backward pass}
\label{appendix:bwd}
Given upstream gradient $\textbf{d}O=\frac{\partial \mathcal{L}}{\partial O}$ for loss $\mathcal{L}$, the goal is to compute $\textbf{d}Q$, $\textbf{d}K$, and $\textbf{d}V$.

For subsequence $i$, the forward pass generates
\begin{equation}
    R_i=\alpha(Q_iK_i^\top)\quad P_i=\text{exp}(R_i)\odot M_i\quad \mathrm{Num}_i=P_iV_i \quad\mathrm{Den}_i=\mathrm{row\_sum}(P_i)
\end{equation}
where $M_i$ is the CQS mask. 

The global merge produces
\begin{equation}
    \mathrm{Num}=\sum_{i\in S}\text{scatter}(\mathrm{Num}_i) \quad \mathrm{Den}=\sum_{i \in S}\text{scatter}(\mathrm{Den}_i)\quad O=\frac{\mathrm{Num}}{\mathrm{Den}}
\end{equation}
The partial derivatives are
\begin{equation}
    \frac{\partial O}{\partial \mathrm{Num}}=\frac{1}{\mathrm{Den}} \quad \quad \frac{\partial O}{\partial \mathrm{Den}}=-\frac{\mathrm{Num}}{\mathrm{Den}^2}
\end{equation}
Below, batch and head dimensions are omitted for simplicity. From the global view,
\begin{equation}
\textbf{d}O[n,d]=\frac{\partial \mathcal{L}}{\partial O[n,d]}
\end{equation}
where $\textbf{d}O[n,d]$ is a scalar from $\textbf{d}O\in \mathbb{R}^{N\times D}$. Then
\begin{equation}
\label{eq:dnum}
    \textbf{d}\mathrm{Num}[n,d]=\frac{\partial \mathcal{L}}{\partial \mathrm{Num}[n]}=\frac{\partial \mathcal{L}}{\partial O[n,d]}\times \frac{\partial O[n,d]}{\partial \mathrm{Num}[n,d]}=\frac{\textbf{d}O[n,d]}{\mathrm{Den}[n]}
\end{equation}
\begin{equation}
\label{eq:dden}
\begin{split}    
    \textbf{d}\mathrm{Den}[n]&=\frac{\partial \mathcal{L}}{\partial \mathrm{Den}[n]}=\frac{\partial \mathcal{L}}{\partial O[n,d]}\times \frac{\partial O[n,d]}{\partial \mathrm{Den}[n]}\\
    &=\sum_d\textbf{d}O[n,d](-\frac{\mathrm{Num}[n,d]}{\mathrm{Den}[n]^2})=-\frac{\sum_d\textbf{d}O[n,d]\mathrm{Num}[n,d]}{\mathrm{Den}[n]^2}
\end{split}
\end{equation}
A compact version of Eqs.~\ref{eq:dnum} and~\ref{eq:dden} is
\begin{equation}
\textbf{d}\mathrm{Num}=\frac{\textbf{d}O}{\mathrm{Den}} \quad \quad \textbf{d}\mathrm{Den}=-\frac{\sum_d\textbf{d}O_d\cdot\mathrm{Num}_d}{\mathrm{Den}^2}
\end{equation}
For subsequence $i$ of length $L$, gathering from global $\textbf{d}\mathrm{Num}\in \mathbb{R}^{N\times D}$ and $\textbf{d}\mathrm{Den}\in \mathbb{R}^{N}$ gives $\textbf{d}\mathrm{Num}_i\in \mathbb{R}^{L\times D}$ and $\textbf{d}\mathrm{Den}_i\in \mathbb{R}^{L}$. The local computations are
\begin{equation}
\label{eq:dvi}
    \textbf{d}V_i[m,d]=\sum_l P_i[l,m]\textbf{d}\mathrm{Num}_i[l,d]
\end{equation}
\begin{equation}
    \textbf{d}P_i[l,m]=\sum_d \textbf{d}\mathrm{Num}_i[l,d]V_i[m,d]+\textbf{d}\mathrm{Den}_i[l]
\end{equation}
\begin{equation}
    \textbf{d}R_i[l,m]=\textbf{d}P_i[l,m]\cdot P_i[l,m]
\end{equation}
\begin{equation}
    \textbf{d}Q_i[l,d]=\alpha \sum_m \textbf{d}R_i[l,m]K_i[m,d]
\end{equation}
\begin{equation}
\label{eq:dki}
    \textbf{d}K_i[m,d]=\alpha \sum_l \textbf{d}R_i[l,m]Q_i[l,d]
\end{equation}
A compact version of Eqs.~\ref{eq:dvi}--\ref{eq:dki} is
\begin{equation}
    \textbf{d}V_i=P_i^\top\textbf{d}\mathrm{Num}_i
\end{equation}
\begin{equation}
\textbf{d}P_i=\textbf{d}\mathrm{Num}_iV_i^\top + \textbf{d}\mathrm{Den}_i\textbf{1}^\top
\end{equation}
\begin{equation}
    \textbf{d}R_i=\textbf{d}P_i\odot P_i
\end{equation}
\begin{equation}
    \textbf{d}Q_i=\alpha\textbf{d}R_iK_i
\end{equation}
\begin{equation}
    \textbf{d}K_i=\alpha\textbf{d}R_i^\top Q_i
\end{equation}
The last step scatters all $\textbf{d}Q_i,\textbf{d}K_i,\textbf{d}V_i \in \mathbb{R}^{L\times D}$ into $\textbf{d}Q,\textbf{d}K,\textbf{d}V \in \mathbb{R}^{N\times D}$.

\section{Supplementary materials}

\begin{algorithm}[htbp]
\caption{CQSA forward pass}
\label{algo:fwd}
\begin{algorithmic}[1]
\Require $Q,K,V \in \mathbb{R}^{B \times H \times N \times D}$, chunk count $c$, decomposition depth $\texttt{itr}$, interest set $\mathcal{I}$, scale $\alpha$
\State $\texttt{subseq\_entries} \gets \textsc{BuildSubseq}(N,c,\texttt{itr},\mathcal{I})$
\State $\mathrm{Num} \gets \mathbf{0}_{B \times H \times N \times D}$
\State $\mathrm{Den} \gets \mathbf{0}_{B \times H \times N}$
\For{each subsequence $i \in \texttt{subseq\_entries}$}
    \State $\texttt{idx} \gets \texttt{token\_ids}[i]$
    \State $Q_i \gets \textsc{Gather}(Q,\texttt{idx})$, $K_i \gets \textsc{Gather}(K,\texttt{idx})$, $V_i \gets \textsc{Gather}(V,\texttt{idx})$
    \State $M_i \gets \texttt{mask}[i]$
    \State $R_i \gets \alpha (Q_i K_i^\top)$
    \State $P_i \gets \exp(R_i) \odot M_i$
    \State $\mathrm{Num}_i \gets P_i V_i$
    \State $\mathrm{Den}_i \gets \textsc{RowSum}(P_i)$
    \State $\mathrm{Num}.\textsc{IndexAdd}(2,\texttt{idx},\mathrm{Num}_i)$
    \State $\mathrm{Den}.\textsc{IndexAdd}(2,\texttt{idx},\mathrm{Den}_i)$
\EndFor
\State $O \gets \mathrm{Num} / \mathrm{Den}.\textsc{Unsqueeze}(-1)$
\State \Return $O$
\end{algorithmic}
\end{algorithm}

\begin{algorithm}[htbp]
\caption{CQSA backward pass. The local attention matrix $P_i$ is recomputed
from $Q_i$, $K_i$.}
\label{algo:bwd}
\begin{algorithmic}[1]
\Require $Q,K,V\in \mathbb{R}^{B \times H \times N \times D}$, upstream $\textbf{d}O$, $\texttt{subseq\_entries}$, $\mathrm{Num}$, $\mathrm{Den}$, scale $\alpha$
\State $\textbf{d}Q \gets \mathbf{0}$, $\textbf{d}K \gets \mathbf{0}$, $\textbf{d}V \gets \mathbf{0}$
\State $\textbf{d}\mathrm{Num} \gets \textbf{d}O / \mathrm{Den}.\textsc{Unsqueeze}(-1)$
\State $\textbf{d}\mathrm{Den} \gets -\textsc{RowSum}_{D}(\textbf{d}O \odot \mathrm{Num}) / (\mathrm{Den} \odot \mathrm{Den})$
\For{each subsequence $i \in \texttt{subseq\_entries}$}
    \State $\texttt{idx} \gets \texttt{token\_ids}[i]$
    \State $Q_i \gets \textsc{Gather}(Q,\texttt{idx})$, $K_i \gets \textsc{Gather}(K,\texttt{idx})$, $V_i \gets \textsc{Gather}(V,\texttt{idx})$
    \State $M_i \gets \texttt{mask}[i]$
    \State $R_i \gets \alpha (Q_i K_i^\top)$
    \State $P_i \gets \exp(R_i) \odot M_i$
    \State $\textbf{d}\mathrm{Num}_i \gets \textsc{Gather}(\textbf{d}\mathrm{Num},\texttt{idx})$, $\textbf{d}\mathrm{Den}_i \gets \textsc{Gather}(\textbf{d}\mathrm{Den},\texttt{idx})$
    \State $\textbf{d}V_i \gets P_i^\top \textbf{d}\mathrm{Num}_i$
    \State $\textbf{d}P_i \gets \textbf{d}\mathrm{Num}_i V_i^\top + \textbf{d}\mathrm{Den}_i \mathbf{1}^\top$
    \State $\textbf{d}R_i \gets \textbf{d}P_i \odot P_i$
    \State $\textbf{d}Q_i \gets \alpha (\textbf{d}R_i K_i)$
    \State $\textbf{d}K_i \gets \alpha (\textbf{d}R_i^\top Q_i)$
    \State $\textbf{d}Q.\textsc{IndexAdd}(2,\texttt{idx},\textbf{d}Q_i)$
    \State $\textbf{d}K.\textsc{IndexAdd}(2,\texttt{idx},\textbf{d}K_i)$
    \State $\textbf{d}V.\textsc{IndexAdd}(2,\texttt{idx},\textbf{d}V_i)$
\EndFor
\State \Return $\textbf{d}Q,\textbf{d}K,\textbf{d}V$
\end{algorithmic}
\end{algorithm}

\begin{figure}[h]
  \centering
  \centerline{\includegraphics[width=0.8\linewidth]{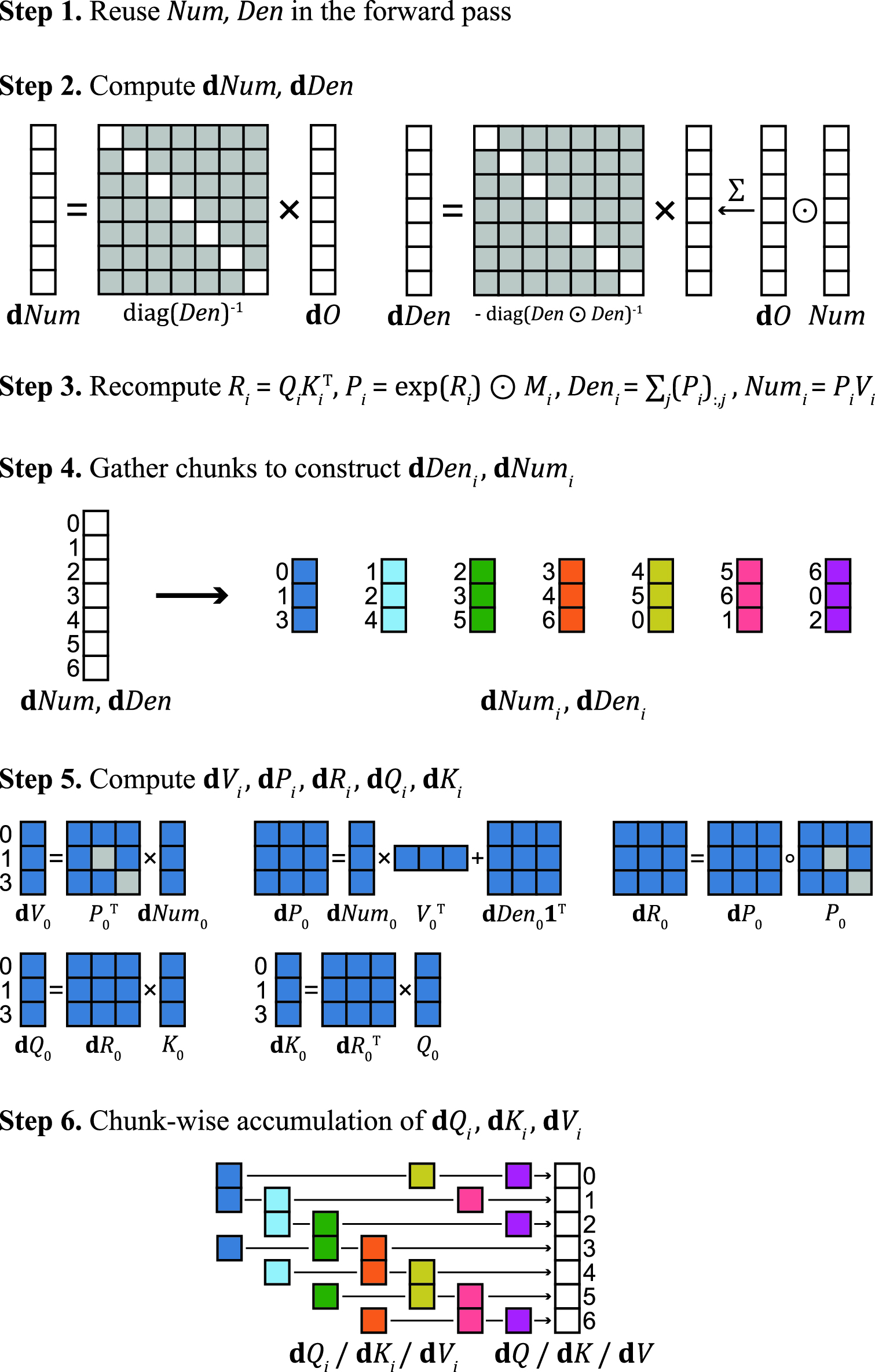}}
  \caption{CQSA backward pass. Step 5 shows only $Seq_0$ because all subsequences follow the same pattern. $\textbf{d}\mathrm{Num} \in \mathbb{R}^{N\times D}$ and $\textbf{d}\mathrm{Den} \in \mathbb{R}^{N\times 1}$.}
  \label{fig:bwd}
\end{figure}

\begin{algorithm}[h]
\caption{\textsc{BuildSubseq}$(N,c,\mathrm{itr},\mathcal{I})$}
\begin{algorithmic}[1]
\Require sequence length $N$, chunk count $c$, decomposition depth $\mathrm{itr}$, interest set $\mathcal{I}$
\Ensure $\texttt{subseq\_entries}$, where each entry has $\texttt{token\_ids}$ and local mask $M_i$
\State $\texttt{subseq\_entries} \gets [\ ]$
\For{each quorum tuple $i=(q_1,\ldots,q_{\mathrm{itr}})\in\{0,\ldots,c-1\}^{\mathrm{itr}}$}
    \State $\texttt{token\_ids} \gets [0,1,\ldots,N-1]$
    \State $\texttt{label\_history} \gets [\ ]$, $\texttt{chunks\_history} \gets [\ ]$
    \For{$t=1$ to $\mathrm{itr}$}
        \State $L \gets |\texttt{token\_ids}|$
        \State $(\texttt{starts},\texttt{ends}) \gets \textsc{BalancedChunkLayout}(L,c)$
        \State $\texttt{chunks} \gets \{(q_t + o)\bmod c \mid o\in\mathcal{I}\}$ (ordered by $\mathcal{I}$)
        \State $\texttt{labels}[\ell] \gets$ chunk id of local index $\ell \in [0,L)$
        \State $\texttt{gather\_idx} \gets \mathrm{Concat}\big([\,\texttt{starts}[u]\ldots\texttt{ends}[u)\ \forall u\in\texttt{chunks}\,\big)$
        \For{$s=1$ to $|\texttt{label\_history}|$}
            \State $\texttt{label\_history}[s] \gets \texttt{label\_history}[s][\texttt{gather\_idx}]$
        \EndFor
        \State append $\texttt{labels}[\texttt{gather\_idx}]$ to $\texttt{label\_history}$
        \State $\texttt{token\_ids} \gets \texttt{token\_ids}[\texttt{gather\_idx}]$
        \State append $(q_t,\texttt{chunks})$ to $\texttt{chunks\_history}$
    \EndFor
    \State $\texttt{group\_runs} \gets [\ ]$
    \For{$t=1$ to $\mathrm{itr}$}
        \State $(\texttt{owner},\texttt{chunks}) \gets \texttt{chunks\_history}[t]$
        \State $\texttt{labels} \gets \texttt{label\_history}[t]$
        \For{each $\texttt{chunk}\in\texttt{chunks}$ with $\texttt{chunk}\neq\texttt{owner}$}
            \State $\texttt{idx} \gets \{\ell \mid \texttt{labels}[\ell]=\texttt{chunk}\}$
            \State $\texttt{runs} \gets \textsc{IndicesToRuns}(\texttt{idx})$
            \If{$\texttt{runs}\neq\emptyset$} append $\texttt{runs}$ to $\texttt{group\_runs}$ \EndIf
        \EndFor
    \EndFor
    \State $\texttt{group\_runs} \gets \textsc{Unique}(\texttt{group\_runs})$
    \State $M_i \gets \textsc{LocalMaskFromGroupRuns}(|\texttt{token\_ids}|,\texttt{group\_runs})$
    \State append $\{\texttt{quorum\_idx}=i,\ \texttt{token\_ids},\ \texttt{mask}=M_i\}$ to $\texttt{subseq\_entries}$
\EndFor
\State \Return $\texttt{subseq\_entries}$
\end{algorithmic}
\end{algorithm}

\begin{table}
\centering\small
\caption{Backward error by gradient tensor ($N=8192$, mean over 10 draws).}
\label{tab:accuracy_grad}
\resizebox{\linewidth}{!}{%
\begin{tabular}{lcccccc}
\toprule
\multirow{2}{*}{Method} & \multicolumn{3}{c}{fp16} & \multicolumn{3}{c}{bf16} \\
\cmidrule(lr){2-4}\cmidrule(lr){5-7}
 & $\mathbf{d}Q$ & $\mathbf{d}K$ & $\mathbf{d}V$ & $\mathbf{d}Q$ & $\mathbf{d}K$ & $\mathbf{d}V$ \\
\midrule
SDPA & $3.075\!\times\!10^{-4}$ & $3.029\!\times\!10^{-4}$ & $2.893\!\times\!10^{-4}$ & $2.458\!\times\!10^{-3}$ & $2.421\!\times\!10^{-3}$ & $2.312\!\times\!10^{-3}$ \\
SDPA (mem-eff.) & $3.739\!\times\!10^{-4}$ & $3.679\!\times\!10^{-4}$ & $2.900\!\times\!10^{-4}$ & $2.980\!\times\!10^{-3}$ & $2.929\!\times\!10^{-3}$ & $2.313\!\times\!10^{-3}$ \\
FlashAttention-2 & $3.075\!\times\!10^{-4}$ & $3.028\!\times\!10^{-4}$ & $2.893\!\times\!10^{-4}$ & $2.458\!\times\!10^{-3}$ & $2.421\!\times\!10^{-3}$ & $2.312\!\times\!10^{-3}$ \\
\midrule
$\mathcal{S}$ \textsc{itr}=0 (undecomposed) & $3.075\!\times\!10^{-4}$ & $3.028\!\times\!10^{-4}$ & $2.893\!\times\!10^{-4}$ & $2.458\!\times\!10^{-3}$ & $2.421\!\times\!10^{-3}$ & $2.312\!\times\!10^{-3}$ \\
$\mathcal{S}$ \textsc{itr}=1 & $3.080\!\times\!10^{-4}$ & $3.024\!\times\!10^{-4}$ & $2.891\!\times\!10^{-4}$ & $2.462\!\times\!10^{-3}$ & $2.424\!\times\!10^{-3}$ & $2.311\!\times\!10^{-3}$ \\
$\mathcal{S}$ \textsc{itr}=2 & $3.085\!\times\!10^{-4}$ & $3.023\!\times\!10^{-4}$ & $2.884\!\times\!10^{-4}$ & $2.465\!\times\!10^{-3}$ & $2.419\!\times\!10^{-3}$ & $2.311\!\times\!10^{-3}$ \\
\bottomrule
\end{tabular}}
\end{table}

\begin{table}
\centering\footnotesize
\setlength{\tabcolsep}{3pt}
\caption{Forward pass, fp16, on one $79.3$ GiB A100. Wall-clock time (s) and peak device memory (GiB, \texttt{max\_memory\_allocated}) are shown from $8192$ to $8\,388\,608$, the longest forward length FlashAttention-2 completes. Parentheses give ratios to FlashAttention-2. These measurements underlie Figure~\ref{fig:overhead_ratio}.}
\label{tab:supp_fwd}
\resizebox{\linewidth}{!}{
\begin{tabular}{lrrrrrrrr}
\toprule
& \multicolumn{4}{c}{Wall-clock time (s)} & \multicolumn{4}{c}{Peak memory (GiB)} \\
\cmidrule(lr){2-5}\cmidrule(lr){6-9}
& & \multicolumn{3}{c}{Stream-CQSA} & & \multicolumn{3}{c}{Stream-CQSA} \\
\cmidrule(lr){3-5}\cmidrule(lr){7-9}
& Flash- & \textsc{itr}=1 & \textsc{itr}=2 & \textsc{itr}=\textsc{auto}$^{*}$
& Flash- & \textsc{itr}=1 & \textsc{itr}=2 & \textsc{itr}=\textsc{auto}$^{*}$ \\
$N$ & Attention-2 & \textsc{acc}=GPU & \textsc{acc}=GPU & \textsc{acc}=CPU
& Attention-2 & \textsc{acc}=GPU & \textsc{acc}=GPU & \textsc{acc}=CPU \\
\midrule
$8\text{K}$ & $0.001$ & $0.034$ $(34.0)$ & $0.121$ $(121)$ & $0.084$ $(84.0)$ & $0.055$ & $0.051$ $(0.93)$ & $0.040$ $(0.73)$ & $0.020$ $(0.36)$ \\
$16\text{K}$ & $0.002$ & $0.051$ $(25.5)$ & $0.161$ $(80.5)$ & $0.163$ $(81.5)$ & $0.102$ & $0.109$ $(1.07)$ & $0.073$ $(0.72)$ & $0.048$ $(0.47)$ \\
$32\text{K}$ & $0.006$ & $0.093$ $(15.5)$ & $0.244$ $(40.7)$ & $0.311$ $(51.8)$ & $0.196$ & $0.210$ $(1.07)$ & $0.138$ $(0.70)$ & $0.089$ $(0.45)$ \\
$64\text{K}$ & $0.027$ & $0.169$ $(6.26)$ & $0.414$ $(15.3)$ & $0.669$ $(24.8)$ & $0.385$ & $0.413$ $(1.07)$ & $0.268$ $(0.70)$ & $0.170$ $(0.44)$ \\
$128\text{K}$ & $0.108$ & $0.335$ $(3.10)$ & $0.730$ $(6.76)$ & $1.34$ $(12.4)$ & $0.762$ & $0.818$ $(1.07)$ & $0.528$ $(0.69)$ & $0.332$ $(0.44)$ \\
$256\text{K}$ & $0.454$ & $0.937$ $(2.06)$ & $1.34$ $(2.95)$ & $3.21$ $(7.08)$ & $1.52$ & $1.63$ $(1.07)$ & $1.05$ $(0.69)$ & $0.656$ $(0.43)$ \\
$512\text{K}$ & $1.86$ & $3.42$ $(1.84)$ & $3.71$ $(2.00)$ & $7.63$ $(4.10)$ & $2.52$ & $3.25$ $(1.29)$ & $2.09$ $(0.83)$ & $1.30$ $(0.52)$ \\
$1\text{M}$ & $7.89$ & $14.2$ $(1.80)$ & $14.8$ $(1.88)$ & $21.2$ $(2.68)$ & $5.04$ & $6.49$ $(1.29)$ & $4.17$ $(0.83)$ & $2.60$ $(0.52)$ \\
$2\text{M}$ & $32.3$ & $52.5$ $(1.62)$ & $65.2$ $(2.02)$ & $66.4$ $(2.06)$ & $10.1$ & $13.0$ $(1.29)$ & $8.34$ $(0.83)$ & $5.19$ $(0.52)$ \\
$4\text{M}$ & $130$ & $203$ $(1.56)$ & $240$ $(1.85)$ & $228$ $(1.76)$ & $20.1$ & $25.9$ $(1.29)$ & $16.7$ $(0.83)$ & $10.4$ $(0.52)$ \\
$8\text{M}$ & $532$ & $802$ $(1.51)$ & $914$ $(1.72)$ & $846$ $(1.59)$ & $40.3$ & $51.8$ $(1.29)$ & $33.3$ $(0.83)$ & $20.7$ $(0.52)$ \\
\bottomrule
\end{tabular}}
\end{table}

\begin{table}
\centering\footnotesize
\setlength{\tabcolsep}{3pt}
\caption{Corresponding forward--backward statistics. As defined in the Metrics paragraph, these measurements time the forward pass followed by gradient computation and therefore represent a full training step.}
\label{tab:supp_bwd}
\resizebox{\linewidth}{!}{
\begin{tabular}{lrrrrrrrr}
\toprule
& \multicolumn{4}{c}{Wall-clock time (s)} & \multicolumn{4}{c}{Peak memory (GiB)} \\
\cmidrule(lr){2-5}\cmidrule(lr){6-9}
& & \multicolumn{3}{c}{Stream-CQSA} & & \multicolumn{3}{c}{Stream-CQSA} \\
\cmidrule(lr){3-5}\cmidrule(lr){7-9}
& Flash- & \textsc{itr}=1 & \textsc{itr}=2 & \textsc{itr}=\textsc{auto}$^{*}$
& Flash- & \textsc{itr}=1 & \textsc{itr}=2 & \textsc{itr}=\textsc{auto}$^{*}$ \\
$N$ & Attention-2 & \textsc{acc}=GPU & \textsc{acc}=GPU & \textsc{acc}=CPU
& Attention-2 & \textsc{acc}=GPU & \textsc{acc}=GPU & \textsc{acc}=CPU \\
\midrule
$8\text{K}$ & $0.002$ & $0.063$ $(31.5)$ & $0.245$ $(122)$ & $0.121$ $(60.5)$ & $0.110$ & $0.107$ $(0.97)$ & $0.082$ $(0.75)$ & $0.060$ $(0.55)$ \\
$16\text{K}$ & $0.006$ & $0.088$ $(14.7)$ & $0.309$ $(51.5)$ & $0.219$ $(36.5)$ & $0.212$ & $0.213$ $(1.00)$ & $0.163$ $(0.77)$ & $0.119$ $(0.56)$ \\
$32\text{K}$ & $0.024$ & $0.154$ $(6.42)$ & $0.429$ $(17.9)$ & $0.406$ $(16.9)$ & $0.416$ & $0.426$ $(1.02)$ & $0.326$ $(0.78)$ & $0.238$ $(0.57)$ \\
$64\text{K}$ & $0.095$ & $0.349$ $(3.67)$ & $0.696$ $(7.33)$ & $0.883$ $(9.29)$ & $0.824$ & $0.852$ $(1.03)$ & $0.652$ $(0.79)$ & $0.475$ $(0.58)$ \\
$128\text{K}$ & $0.380$ & $1.03$ $(2.71)$ & $1.39$ $(3.65)$ & $3.08$ $(8.11)$ & $1.64$ & $1.71$ $(1.04)$ & $1.30$ $(0.79)$ & $0.950$ $(0.58)$ \\
$256\text{K}$ & $1.57$ & $4.56$ $(2.91)$ & $5.00$ $(3.19)$ & $7.36$ $(4.70)$ & $3.27$ & $3.41$ $(1.04)$ & $2.61$ $(0.80)$ & $1.90$ $(0.58)$ \\
$512\text{K}$ & $6.20$ & $14.8$ $(2.39)$ & $16.4$ $(2.64)$ & $20.6$ $(3.32)$ & $5.04$ & $6.82$ $(1.35)$ & $5.22$ $(1.04)$ & $3.80$ $(0.75)$ \\
$1\text{M}$ & $25.3$ & $56.2$ $(2.22)$ & $65.5$ $(2.59)$ & $66.0$ $(2.61)$ & $10.1$ & $13.6$ $(1.35)$ & $10.4$ $(1.04)$ & $7.60$ $(0.75)$ \\
$2\text{M}$ & $101$ & $216$ $(2.13)$ & $256$ $(2.52)$ & $235$ $(2.31)$ & $20.1$ & $27.3$ $(1.35)$ & $20.9$ $(1.04)$ & $15.2$ $(0.76)$ \\
$4\text{M}$ & $407$ & $847$ $(2.08)$ & $983$ $(2.41)$ & $885$ $(2.17)$ & $40.3$ & $54.5$ $(1.35)$ & $41.7$ $(1.04)$ & $30.4$ $(0.76)$ \\
\bottomrule
\end{tabular}}
\end{table}

\begin{figure}
\centering
\includegraphics[width=\linewidth]{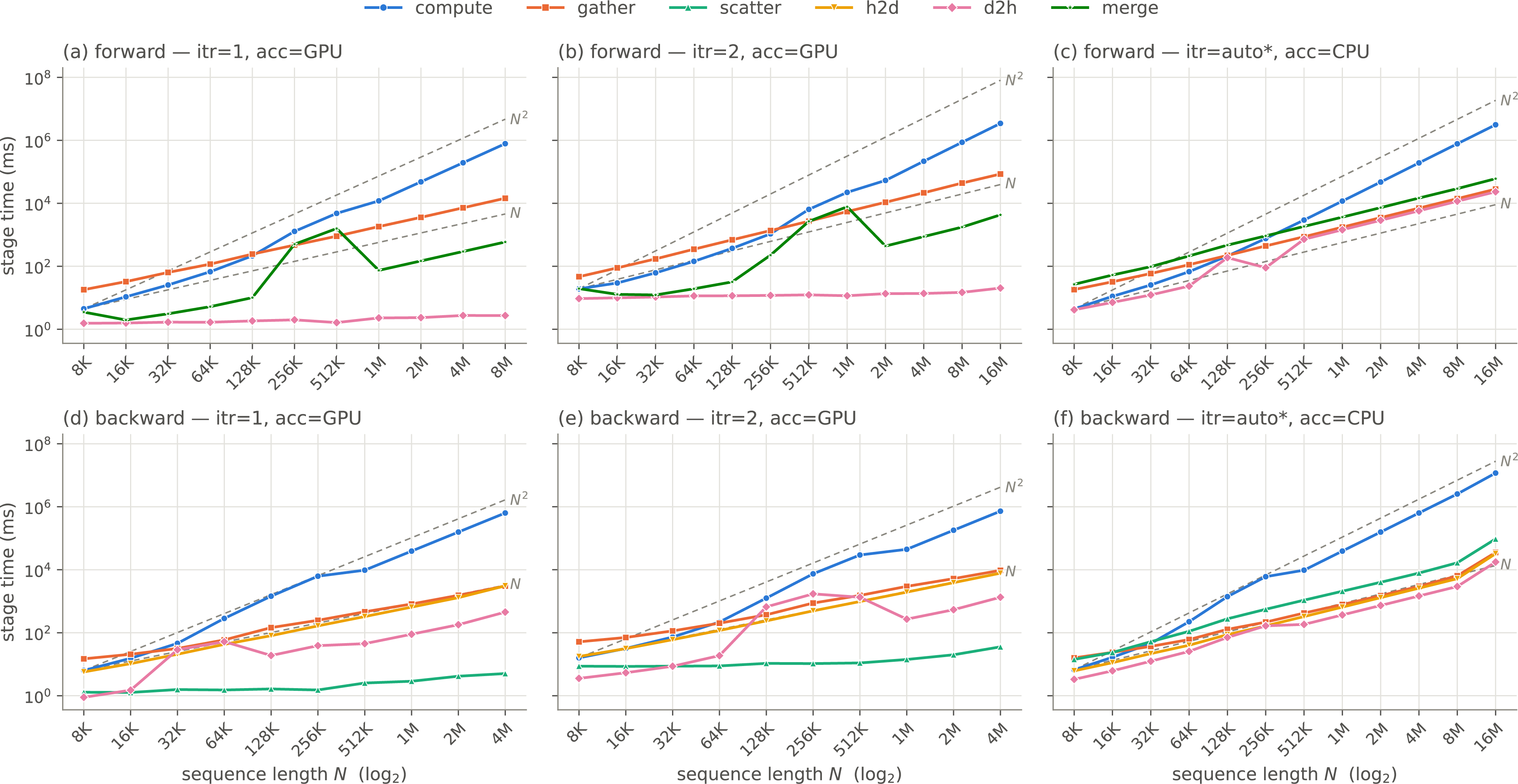}
\caption{The measurements of Figure~\ref{fig:stages} in absolute time on log-log axes. Grey guides of slope one and two are anchored to the local kernel at the shortest length to mark linear and quadratic growth in $N$. They are references, not fits. Time queued behind another stream is excluded because it runs concurrently with the charged stage. As in Figure~\ref{fig:stages}, backward panels cover the backward pass alone.}

\label{fig:supp_stages_abs}
\end{figure}

\end{document}